%% file: 2021_jmlr_qanta.tex
\documentclass[twoside,11pt]{article}

\usepackage{style/jmlr2e}

\ShortHeadings{Quizbowl}{Rodriguez, Feng, Iyyer, He, and Boyd-Graber}
\firstpageno{1}

\usepackage[utf8]{inputenc} 
\usepackage[T1]{fontenc}    
\usepackage{hyperref}       
\usepackage{url}            
\usepackage{booktabs}       
\usepackage{amsfonts}       
\usepackage{nicefrac}       
\usepackage{microtype}      
\usepackage{dsfont}
\usepackage{xcolor}
\usepackage{caption}
\usepackage{subcaption}
\usepackage{enumitem}
\usepackage{bbm}
\usepackage{tabularx}
\usepackage{hhline}
\usepackage{makecell}
\usepackage[toc,page]{appendix}
\usepackage{enumitem}
\usepackage{etoolbox}
\usepackage{color,soul}
\usepackage{tikz}
\usepackage{textcomp}
\usepackage{etoolbox}
\usetikzlibrary{shapes.geometric}

\makeatletter
\patchcmd{\SOUL@ulunderline}{\dimen@}{\SOUL@dimen}{}{}
\patchcmd{\SOUL@ulunderline}{\dimen@}{\SOUL@dimen}{}{}
\patchcmd{\SOUL@ulunderline}{\dimen@}{\SOUL@dimen}{}{}
\newdimen\SOUL@dimen
\makeatother

\graphicspath{{2021_jmlr_qanta/commit_fig/}{2021_jmlr_qanta/auto_fig/}{2021_jmlr_qanta/figures/}}

\newif\ifcomment\commenttrue
\input{style/preamble}

\newcommand{\dan}[0]{\abr{dan}}

\newcommand{\gru}[0]{\abr{gru}}

\newcommand{\nquestions}[0]{119,093}
\newcommand{\ntotalquestions}[0]{132,849}

\newcommand{\wikidumpdate}[0]{4/18/2018}
\newcommand{\simplequestions}[0]{SimpleQuestions}

\newcommand{\R}{\mathbb{R}}
\definecolor{xred}{HTML}{DB4437}
\definecolor{xyellow}{HTML}{F4B400}
\definecolor{xgreen}{HTML}{0F9D58}

\def\genbox#1#2#3#4#5#6{
    \leavevmode\raise#4bp\hbox to#5bp{\vrule height#5bp depth0bp width0bp
    \pdfliteral{q .5 w \csname #2COLOR\endcsname\space RG
                       \csname #3PDF\endcsname{#5}{#6} S Q
             \ifx1#1 q \csname #2COLOR\endcsname\space rg 
                       \csname #3PDF\endcsname{#5}{#6} f Q\fi}\hss}}

\def\sqbox      #1#2{\genbox{#1}{#2}  {sq}       {0}   {4.5}  {2.25}}
\def\trianbox   #1#2{\genbox{#1}{#2}  {trian}    {0}   {5}    {2.5}}
\def\uptrianbox #1#2{\genbox{#1}{#2}  {uptrian}  {0}   {5}    {2.5}}
\def\circbox    #1#2{\genbox{#1}{#2}  {circ}     {0}   {5}    {2.5}}
\def\diabox     #1#2{\genbox{#1}{#2}  {dia}      {-.5} {6}    {3}}

\newcommand*{\tcircle}[1]{\circbox1{#1}}
\newcommand*{\tsquare}[1]{\sqbox1{#1}}
\newcommand*{\tdiamond}[1]{\uptrianbox1{#1}}
\newcommand*{\ttriangle}[1]{\trianbox1{#1}}

\definecolor{qbquestion}{HTML}{9EE6FF}
\renewrobustcmd{\bfseries}{\fontseries{b}\selectfont}
\renewrobustcmd{\boldmath}{}

\setitemize{noitemsep,topsep=0pt,parsep=0pt,partopsep=0pt}

\begin{document}

\title{Quizbowl: The Case for Incremental Question Answering}

\author{\name Pedro Rodriguez\email pedro@cs.umd.edu\\
	\name Shi Feng\email shifeng@cs.umd.edu\\
	\addr Department of Computer Science\\
	University of Maryland at College Park\\
	College Park, MD
	\AND
	\name Mohit Iyyer\email miyyer@cs.umass.edu\\
	\addr College of Information and Computer Sciences\\
	University of Massachusetts Amherst\\
	Amherst, MA
	\AND
	\name He He\email hehe@cs.nyu.edu\\
	\addr Department of Computer Science, Courant Institute\\
	New York University\\
	New York, NY
	\AND
	\name Jordan Boyd-Graber\email jbg@umiacs.umd.edu \\
	\addr Department of Computer Science, iSchool, \abr{umiacs}, \abr{lsc} \\
	University of Maryland at College Park\\
	College Park, MD}

\editor{}

\maketitle

\begin{abstract}%
	\input{2021_jmlr_qanta/sections/00-abstract.tex}
\end{abstract}


\input{2021_jmlr_qanta/sections/10-intro.tex}

\input{2021_jmlr_qanta/sections/20-task.tex}
\input{2021_jmlr_qanta/sections/30-dataset.tex}
\input{2021_jmlr_qanta/sections/40-system.tex}
\input{2021_jmlr_qanta/sections/50-guessing.tex}
\input{2021_jmlr_qanta/sections/60-buzzing.tex}

\input{2021_jmlr_qanta/sections/65-eval.tex}
\input{2021_jmlr_qanta/sections/70-live.tex}
\input{2021_jmlr_qanta/sections/80-rel.tex}
\input{2021_jmlr_qanta/sections/85-future.tex}

\input{2021_jmlr_qanta/sections/90-conc.tex}

\acks{}
Rodriguez and Boyd-Graber are supported by National Science Foundation Grants IIS1320538 and IIS1822494.
Feng is supported under subcontract to Raytheon BBN Technologies by DARPA award HR001-15-C-0113.
Amazon Web Services Cloud Credits for Research provided computational resources that supported many of our experiments and internal infrastructure.
Any opinions, findings, conclusions, or recommendations expressed here are those of the authors and do not necessarily reflect the view of the sponsor.

Many individuals contributed their ideas and work to this paper.
Davis Yoshida was influential in developing a Wikipedia data augmentation method (eventually subsumed by \bert{}) and contributed to many insightful discussions.
Others who have been intellectually involved in our work with \qb{} include Brianna Satinoff, Anupam Guha, Danny Bouman, Varun Manjunatha, Hal Daume III, Leonardo Claudino, and Richard Socher.
We also thank the members of the \abr{clip} lab at the University of Maryland for helpful discussion.

Next, we appreciate those that improved this paper through comments, edits, and reference suggestions on the manuscript.
In particular, we thank Yogarshi Vyas, Joseph Barrow, Alvin Grissom, Hal Daum\'e III, Philip Resnik, and Kianté Brantley for their feedback.

Nathan Murphy and R. Robert Hentzel have been incredibly supportive of using \qb{} for outreach with the public; they have multiple times hosted our exhibition events at the high school national championship tournaments.
We also thank the participants that played against machine systems in these events; these include Ophir Lifshitz, Auroni Gupta, Jennie Yang, Vincent Doehr, Rahul Keyal, Kion You, James Malouf, Rob Carson, Scott Blish, Dylan Minarik, Niki Peters, Colby Burnett, Ben Ingram, Alex Jacob, and Kristin Sausville.
Ken Jennings played against our system at an event hosted by Noah Smith at the University of Washington.
Ikuya Yamada and Studio \abr{ousia} entered their systems which competed against human teams at several of our events.

Finally, this work would never have been possible without the support
of the \qb{} community.
We are grateful to the original authors of questions, and those who
helped us collect our current dataset.
The maintainers of \url{quizdb.org} and \url{protobowl.com} allowed us
to use their websites to build our dataset.
The first versions of work in \qb{} used a dataset collected by
Shivaram Venkataraman and Jerry Vinokurov. These first versions used questions from the \qb{} Academic Competition Federation.
National Academic Quiz Tournaments, LLC provided access to their
proprietary questions which we used in prior iterations of our
systems.

\appendix
\begin{appendices}
	\input{2021_jmlr_qanta/sections/appendix.tex}

\end{appendices}

\bibliography{bib/journal-full,bib/qanta}

\end{document}

%% file: style/preamble.tex
\usepackage[a-1b]{pdfx}

\usepackage{framed}
\usepackage{mdwlist}
\usepackage{latexsym}
\usepackage{colortbl}
\usepackage{xcolor}
\usepackage{nicefrac}
\usepackage{booktabs}
\usepackage{amsfonts}
\usepackage[T1]{fontenc}
\usepackage{bold-extra}
\usepackage{amsmath}
\usepackage{amssymb}
\usepackage{bm}
\usepackage{graphicx}
\usepackage{mathtools}
\usepackage{microtype}
\usepackage{multirow}
\usepackage{multicol}
\usepackage{todonotes}
\usepackage{latexsym,comment}
\usepackage[normalem]{ulem}

\newcommand*{\missingreference}{{\Huge \colorbox{red}{?reference?}}}
\newcommand*{\missingcitation}{{\Huge \colorbox{red}{?citation?}}}
\makeatletter
\def\@setref#1#2#3{%
    \ifx#1\relax
        \protect\G@refundefinedtrue
        \nfss@text{\reset@font\missingreference}%
        \@latex@warning{Reference `#3' on page \thepage \space
            undefined}%
    \else
        \expandafter#2#1\null
    \fi}
\def\@citex[#1]#2{\leavevmode
    \let\@citea\@empty
    \@cite{\@for\@citeb:=#2\do
        {\@citea\def\@citea{,\penalty\@m\ }%
            \edef\@citeb{\expandafter\@firstofone\@citeb\@empty}%
            \if@filesw\immediate\write\@auxout{\string\citation{\@citeb}}\fi
            \@ifundefined{b@\@citeb}{\hbox{\reset@font\missingcitation}%
                \G@refundefinedtrue
                \@latex@warning
                {Citation `\@citeb' on page \thepage \space undefined}}%
            {\@cite@ofmt{\csname b@\@citeb\endcsname}}}}{#1}}
\makeatother

\newcommand{\gem}[1]{\mbox{\textsc{gem}}}
\newcommand{\abr}[1]{\textsc{#1}}
\newcommand{\camelabr}[2]{{\small #1}{\textsc{#2}}}

\DeclareMathOperator*{\argmax}{arg\,max}

\newcommand{\e}[2]{\mathbb{E}_{#1}\left[ #2 \right] }

\newcommand{\ind}[1]{\mathds{1}\left[ #1 \right] }

\newcommand{\hidetext}[1]{}
\newcommand{\ignore}[1]{}

\ifcomment
    \newcommand{\pinaforecomment}[3]{\colorbox{#1}{\parbox{.8\linewidth}{#2: #3}}}
\else
    \newcommand{\pinaforecomment}[3]{}
\fi

\newcommand{\smallurl}[1]{ \begin{tiny}\url{#1}\end{tiny}}

\definecolor{lightblue}{HTML}{3cc7ea}
\definecolor{CUgold}{HTML}{CFB87C}
\definecolor{grey}{rgb}{0.95,0.95,0.95}
\definecolor{ceil}{rgb}{0.57, 0.63, 0.81}
\definecolor{UMDred}{HTML}{ed1c24}
\definecolor{UMDyellow}{HTML}{ffc20e}

\newcommand{\qb}[0]{\abr{qb}}
\newcommand{\qa}[0]{\abr{qa}}
\newcommand{\nlp}[0]{\abr{nlp}}

\newcommand{\triviaqa}{\camelabr{Trivia}{qa}}
\newcommand{\searchqa}{\camelabr{Search}{qa}}

\newcommand{\newsqa}{\camelabr{News}{qa}}
\newcommand{\trecqa}{\camelabr{Trec}{qa}}

\newcommand{\qanta}{\textsc{qanta}}

\newcommand{\squad}{\textsc{sq}{\small u}\textsc{ad}}
\newcommand{\nq}[0]{\abr{nq}}

\newcommand{\glove}{\abr{gl}{\small o}\abr{ve}}

\newcommand{\bert}{\abr{bert}}

%% file: 2021_jmlr_qanta/sections/00-abstract.tex
Scholastic trivia competitions test knowledge and intelligence through mastery of question answering.
Modern question answering benchmarks are one variant of the Turing test.
Specifically, answering a set of questions as well as a human is a minimum bar towards demonstrating human-like intelligence.
This paper makes the case that the format of one competition---where participants can answer in the middle of hearing a question (incremental)---better differentiates the skill between (human or machine) players.
Additionally, merging a sequential decision-making sub-task with question answering provides a good setting for research in model calibration and opponent modeling.
Thus, embedded in this task are three machine learning challenges: (1) factoid \qa{} over thousands of Wikipedia-like answers, (2) calibration of the \qa{} model's confidence scores, and (3) sequential decision-making that incorporates knowledge of the \qa{} model, its calibration, and what the opponent may do.
We make two contributions: (1) collecting and curating a large factoid \abr{qa} dataset and an accompanying gameplay dataset, and (2) developing a model that addresses these three machine learning challenges.
In addition to offline evaluation, we pitted our model against some of the most accomplished trivia players in the world in a series of exhibition matches spanning several years.
Throughout this paper, we show that collaborations with the vibrant trivia community have contributed to the quality of our dataset, spawned new research directions, and doubled as an exciting way to engage the public with research in machine learning and natural language processing.


%% file: 2021_jmlr_qanta/sections/10-intro.tex
\section{Introduction}
\label{sec:intro}

\begin{figure}[ht]
  \begin{center}
    \tikz\node[draw=black!40!lightblue,inner sep=1pt,line width=0.3mm,rounded corners=0.1cm]{ 
      \begin{tabular}{p{.9\textwidth}}
        At its premiere, the librettist of this opera portrayed a character who asks
        for a glass of wine with his dying wish. That character in this opera is
        instructed to ring some bells to summon his love. At its beginning, a man who
        claims to have killed a serpent has a padlock put on his mouth because of
        his lying. The plot of this opera concerns a series of tests that Tamino must
        undergo to rescue Tamina from Sorastro. For 10 points, name this Wolfgang
        Mozart opera titled for an enchanted woodwind instrument. \\
        \textbf{Answer:} \underline{The Magic Flute}
      \end{tabular}
    };
  \end{center}
  \caption{ \qb{} is a trivia game where questions begin with clues
    that are initially difficult, but become progressively easier
    until a giveaway at the end of the question.  Players answer as
    soon as they know the answer so as a result the earlier they
    answer the more knowledgeable they are.  For example, answering
    after the first sentence indicates the player recognizes the
    librettist (Emanual Schikaneder) and knows that they played
    Papageno in \textit{The Magic Flute} (die Zauberfl\"ote). 
    Answering at the end of the question only
    requires surface knowledge of Mozart's opera works.  }
  \label{fig:ex}
\end{figure}

Answering questions is an important skill for both humans and
computers.
Exams form the foundation of educational systems and---for many
societies---of the civil system~\citep{Fukuyama-95}.
Computers answering questions in the Turing test is the standard
definition of artificial intelligence~\citep{Turing-95}.
But another more trivial form of question answering is more pervasive
in popular culture.

Trivia games are pervasive and popular: from quizzing in
India~\citep{roy-16} to ``What? Where? When?'' in
Russia~\citep{korin-02} to ``Who wants to be a Millionaire''~\citep{clarke2001million,lam2003million}, trivia encourages people to acquire, recall,
and reason over facts.
For computers, \citet{Yampolskiy-13} argues that these skills are
\abr{ai}-complete: solve question answering and you have
solved \abr{ai} generally.
Our central thesis in this article is that the intense research in
question answering would benefit in adopting the innovations and
lessons learned from human trivia competition, as embodied in a trivia
format called \textbf{Quizbowl} (\qb{}).

In \qb{}, questions are posed \emph{incrementally}---word by
word---and players must \emph{interrupt} the question when they know
the answer (Figure~\ref{fig:ex}).
Thus, it rewards players who can answer with less information than
their opponents.
This is not just a gimmick to separate it from other question
answering formats: players must simultaneously think about what is the
most likely answer and after every word decide whether it is better to
answer or wait for more information.
To succeed, players and machines alike must answer questions, maintain
accurate estimates of their confidence, and factor their opponents'
abilities.
The combination of these skills makes \qb{} challenging for machine
learning algorithms.

A dedicated and skilled community forged \qb{} over decades
(Section~\ref{sec:task}), creating a diverse and large dataset
(Section~\ref{sec:datasets}).
We refer to this dataset as the \qanta{} dataset because (in our
opinion) \textbf{Q}uestion \textbf{A}nswering is \textbf{N}ot a
\textbf{T}rivial \textbf{A}ctivity.\footnote{
  Dataset available at
  \url{http://datasets.qanta.org}.}

Playing \qb{} requires deciding \emph{what} to answer
(Section~\ref{sec:guessing}) and \emph{when} to answer
(Section~\ref{sec:buzzing}).
Our final contribution is a framework that combines independent
systems for each of these sub-tasks.
Despite its simplicity, our implementation of this framework is
competitive with the best players.
Section~\ref{sec:live} showcases \qb{} as a platform for
simultaneously advancing research and educating the public about the
limits of machine learning through live human--computer competitions.
Finally, we discuss ongoing and future research using trivia questions
to build machines that are as capable of reasoning and connecting
facts as humans.

%% file: 2021_jmlr_qanta/sections/20-task.tex


\section{Why Quizbowl?}
\label{sec:task}

When discussing machine learning and trivia, the elephant in the room
is always \abr{ibm}'s tour-de-force
match~\citep{Ferrucci2010BuildingWA} against Ken Jennings and Brad
Rutter on \textit{Jeopardy!}
Rather than ignore the obvious comparisons, we take this on directly
and use the well-known \textit{Jeopardy!} context---which we gratefully
acknowledge as making our own work possible---as a point of
comparison, as
\qb{} is a better
differentiator of skill between participants, be they human or machine
(Sections~\ref{sec:race}
and~\ref{sec:pyramid}).\footnote{\citet{Boyd-Graber:Satinoff:He:Daume-III-2012}
    introduce \qb{} as a factoid question answering task,
    \citet{Iyyer:Manjunatha:Boyd-Graber:Daume-III-2015} further develop
    algorithms for answering questions, and \citet{he2016opponent}
    improve live play.
    This journal article synthesizes our prior work
    scattered across disparate publications, drops artificial
    limitations (e.g., ignoring categories or rare answers), and
    evaluates models in offline, online, and live environments.
    Moreover, it connects the earlier work with question answering
    datasets that followed such as \squad{}.}
While this section will have more discussion of the history of trivia
than the typical machine learning paper, the hard-won lessons humans
learned about question answering transfer into machine question
answering.

\begin{figure}
    \centering
    \includegraphics[width=0.8\linewidth]{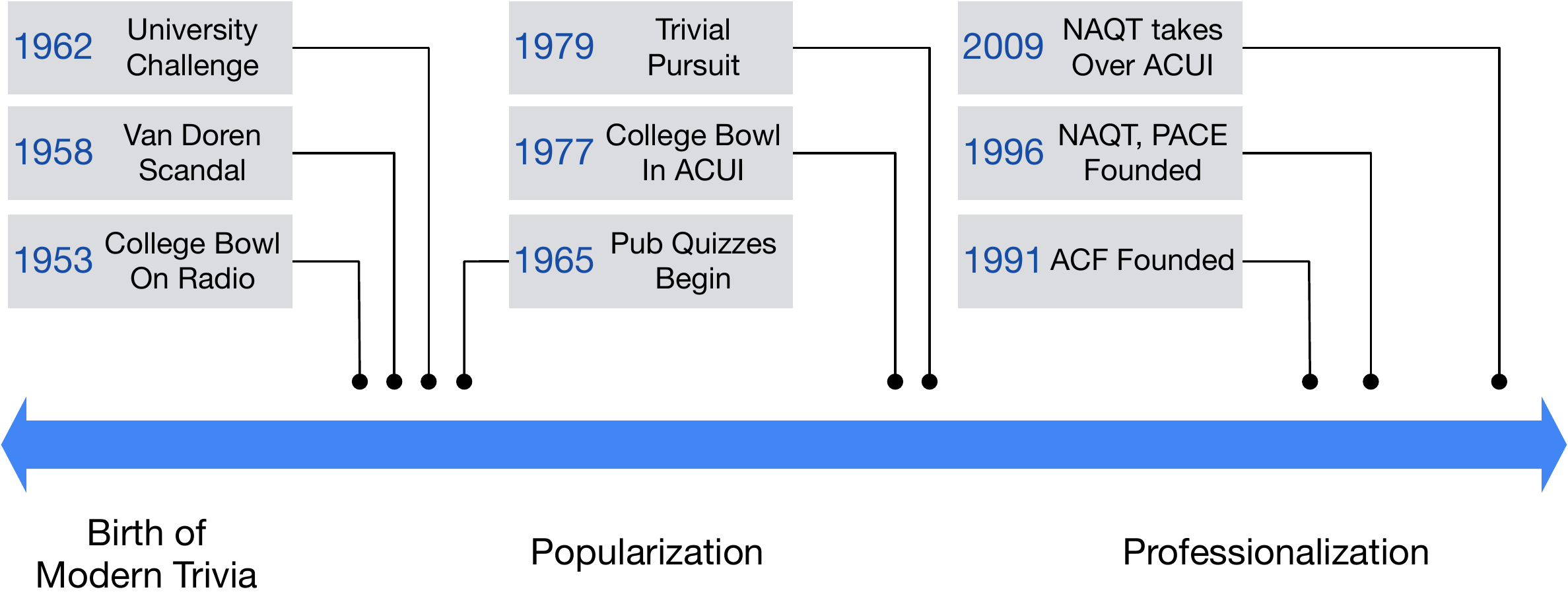}
    \caption{Trivia has gone from a laid-back pastime to an organized,
        semi-professional competition format.  The \qb{} framework, in
        particular, which arose from College Bowl (\abr{us}) and
        University Challenge (\abr{uk}) emphasizes fairness and the
        ability to discover the better question answerer.  As
        organizations such as the Academic Competition Federation and
        National Academic Quiz Tournaments emerged, the format has focused
        on academic, well-run tournaments.}
    \label{fig:history}
\end{figure}

The \qa{} format categorization of \citet{gardner2020format} names three tasks where framing the problem as \qa{} is \emph{useful}: (1) filling human information needs, (2) \qa{} as annotation or probe, and (3) as a transfer mechanism.
Like \searchqa{}~\citep[Jeopardy!]{Dunn2017SearchQAAN}, \qb{} does not
explicitly probe specific linguistic phenomena; it uses language to
ask what humans know.
In contrast to questions posed to search engines or digital
assistants~\citep{Nguyen2016MSMA,nq19},
\qb{} is less ambiguous: question writers ensure that the descriptions
uniquely identify one and only one answer, a non-trivial goal~\citep{Voorhees2000BuildingAQ}.
Thus, the goals and challenges in \qb{} are similar to---yet
distinct from---open domain information-seeking.

The \qb{} format is compelling and consistent because of its evolution (Figure~\ref{fig:history}) over
its fifty-year history.\footnote{ After returning from World War II
    and inspired by \abr{uso} morale-building activities, Canadian Don
    Reid sketched out the format with the first host Allen
    Ludden.
    After a radio premiere in 1953, \textit{College Bowl} moved to
    television in 1959 and became the first television show to win a
    Peabody~\cite{Baber-15}.
    The format established many careers: the future president of the
    National Broadcasting Corporation (\abr{nbc}), Grant Tinker,
    served as the game's first scorekeeper (the newly designed game and its scoring was
    so confusing that Allen Ludden often had to \textit{ad lib} to let
    Tinker catch up).
    The format was intriguing enough that Granada studios copied
    it---initially without permission---into what became the \abr{uk}
    cultural touchstone \textit{University
        Challenge}~\citep{taylor_mcnulty_meek}, establishing the career of
    Bamber Gascoigne.}
Many of the challenges the \abr{nlp} community
faces in collecting good question answering datasets at scale~\citep{hermann2015teaching} were first encountered by trivia aficionados. For example,
avoiding predictive yet useless patterns in data~\citep{Jia2017AdversarialEF,DBLP:conf/emnlp/KaushikL18}; players do not like re-used clues making questions trivially easy. Trivia players also aim to write questions that require multi-hop reasoning; datasets like HotPotQA~\citep{Yang2018HotpotQAAD} have similar goals, but writing questions that truly require multi-hop reasoning is challenging~\citep{Min2019CompositionalQD}.
We distill these lessons, describe the craft of question writing that
makes \qb{} a compelling question answering task
(Section~\ref{sec:craft}), and enumerate some \abr{nlp} challenges
required to truly solve \qb{} (Section~\ref{sec:challenges}).
We conclude by framing \qb{} as a hybrid task between question
answering and sequential decision-making
(Section~\ref{sec:form-task}).

\subsection{What is a Buzzer Race?}
\label{sec:race}

The scapegoat for every \textit{Jeopardy!} loser and the foundation of every
Jeopardy! winner is the {\bf buzzer}~\citep{trebekistan}.
A buzzer is a small handheld device that players press to signal that
they can correctly respond to a clue.
The fundamental difference between \textit{Jeopardy!} and \qb{}---and what
makes \qb{} more suitable for research---is how clues are revealed and
how players use the buzzer.

Jeopardy! is a television show and uses the buzzer to introduce
uncertainty, randomness, and thus excitement for the viewer at home.
In \textit{Jeopardy!}, players can only use the buzzer when the moderator has
finished reading the question.\footnote{In \textit{Jeopardy!} terminology is
    reversed so that a moderator reads clues termed \emph{answers} to
    which players must supply the correct \emph{question}.  To avoid
    confusion, we follow standard terminology.}
If players use the buzzer before the question is finished,
they are locked out and prevented from answering the question for a
fraction of a second (an eternity in the fast-paced game of
Jeopardy!).

This advantaged Watson in its match against
two opponents with feeble human thumbs and reflexes, as
Jeopardy! uses the buzzer to determine who among those who know the
answer \emph{has the fastest reflexes}.\footnote{In a Ken Jennings
    interview with \abr{npr}~\cite{malone-19}, the host Kenny Malone
    summarized it well as ``To some degree, \textit{Jeopardy!} is kind
    of a video game, and a crappy video game where it's, like, light
    goes on, press button---that's it.''  Ken Jennings agreed, but
    characterized it as ``beautiful art and not a really crappy video
    game''.}
While Watson gets an electronic signal when it was allowed to buzz,
the two humans watch for a light next to the \textit{Jeopardy!} game board to
know when to buzz.
Thus, Watson---an electronic buzzing machine---snags the first choice
of questions, while the two humans fight over the scraps.
In \textit{Jeopardy!} reflexes are almost as important as knowledge.
Next we show how the structure of \qb{} questions and its use of a
buzzer rewards depth of knowledge rather than reflexes.

\subsection{Pyramidality and Buzzers}
\label{sec:pyramid}

In contrast, \qb{} is a game honed by trivia enthusiasts which uses
buzzers as a tool to determine \emph{who knows the most about a
    subject}.
This is possible because the questions are \emph{interruptable}.
Unlike \textit{Jeopardy!}, players can interrupt the questions when they know
the answer (recall questions are multi-sentence in \qb{}).
This would make for bad television (people like to play along at home
and cannot when they cannot hear the whole question), but makes for a
better trivia game that also requires decision-making under uncertainty.

This alone is insufficient however; if an easy clue appears early in
the question then knowing hard clues later in the question is
irrelevant.
Questions that can be answered with only a fraction of their input are
a bad foundation for research~\citep{Sugawara2018WhatMR,Feng2019-tc}.
\qb{} addresses this problem by structuring questions
\emph{pyramidally}.
In pyramidal questions, clues are incorporated so that harder, more
obscure information comes first in the question, and easier, more
obvious information comes at the end of the question.
Thus, when a player answers before their opponents,
they are more knowledgeable than their opponents.

This also makes \qb{} an attractive machine learning research domain.
The giveaways are often easy for computers too: they are prominent on
Wikipedia pages and have appeared in many questions.
Thus, it is easy for computers to answer most questions \emph{at some
    point}: \qb{} is not an impossibly difficult problem.
The challenge then becomes to answer the questions \emph{earlier},
using more obscure information and higher-order reasoning.

Humans who play \qb{} have the same yearning; they can answer most of
the questions, but they want to deepen their knowledge to buzz in just a
little earlier.
They keep practicing, playing questions and going to tournaments to
slowly build skill and knowledge.
\qb{} is engineered for this to be a rewarding experience.

The same striving can motivate researchers: it does not take much to
buzz in a word earlier.
As small incremental improvements accumulate,
we can have more robust, comprehensive question answering systems.
And because \qb{} has a consistent evaluation framework, it is easy to
see whose hard work has paid off.

Thus, the form of \qb{} questions---the product of decades of refining
how to measure the processing and retrieval of information of
humans---can also compare machines' question answering ability.
We next describe the cultural norms of question writing in the \qb{}
community that contribute to making it a challenging task for humans
and machines alike.

\subsection{The Craft of Question Writing}
\label{sec:craft}

The goal of \qb{} is to reward ``real'' knowledge.
This goal is the product of a long history that has resulted in
community norms that have evolved the competition into a thriving,
carefully designed trivia ecosystem.
By adopting these conventions, machine learning can benefit from the
best practices for question answering evaluation without repeating the
same mistakes.

Every year, question writers in the community focus on creating high
quality questions that are novel and pyramidal.
Experts write thousands of questions each year.\footnote{Regional
    competition questions are written by participants; championship
    competition questions are written by professionals hired by either
    the Academic Competition Federation (\abr{acf}), National Academic
    Quiz Tournaments (\abr{naqt}), or the Partnership for Academic
    Competition Excellence (\abr{pace}).  While the exact organizational
    structure varies, initial draft questions are vetted and edited by domain
    experts.}
To maintain the quality and integrity of competition, the community
enforces rules consistent with machine learning's question for
generalization as described by \citet{boydgraber2020nerds}: avoiding
ambiguity, ensuring correctness, eschewing previously used clues, and allowing for fair comparisons
between teams~\citep{write-qb-ucb,write-qb-forum,write-qb-subash} of 10,000
middle school students, 40,000 high school students, and 3,200 college
students~\citep{naqt-participation}.
At the same time, in preparation for tournaments students study questions from previous years.

These dueling groups---players and writers---create a factual arms race that is the foundation for the quality of \qb{} questions.
Aligning annotators'
motivations~\citep{ahn-06}---such as playing a game---with the goals of the data
collection improves the quality and quantity of data.
A similar arms race between dataset exploiters (attackers) and those
seeking to make datasets more robust (defenders) exists in other
machine learning domains like computer
vision~\citep{carlini2017attacks,hendrycks2019natural} and Build-It,
Break-It, (Fix-It) style
tasks~\citep{ettinger2017build,thorne2019fever2,dinan2019build,nie2019adversarial}.

In \qb{}, answers are uniquely identifiable named entities such as---but not limited to---people, places, events, and literary works.
These answers are ``typified by a noun phrase'' as in \citet{Kupiec1993MURAXAR} and later in the \abr{trec} \qa{} track~\citep{voorhees2003evaluating}.
Similar answer types are also used by other factoid question answering
datasets such as
SimpleQuestions~\citep{DBLP:journals/corr/BordesUCW15},
SearchQA~\citep{Dunn2017SearchQAAN},
TriviaQA~\citep{JoshiTriviaQA2017}, and
NaturalQuestions' short answers~\citep{nq19}.
In its full generality, \qb{} is an Open Domain \qa{} task~\citep{chen2017reading,chen2020odqa}.
However, since the vast majority of answers correspond to one of the six million entities in Wikipedia (Section~\ref{sec:q-proc}),\footnote{
    A minority of answers cannot be mapped.
    Some answers do not have a page because Wikipedia is incomplete
    (e.g., not all book characters have Wikipedia pages).
    Other entities are excluded by Wikipedia editorial decisions: they
    lack notability, are combined with other entities(e.g., \underline{Gargantua and
        Pantagruel} and \underline{Romulus and Remus}).
    Other abstract answers will likely never have Wikipedia pages (\underline{women with one leg}, \underline{ways Sean Bean has died
        in films}).
} we approximate the open-domain setting by defining this as our
source of answers (Section~\ref{sec:ds-tasks} reframes this in reading
comprehension's  span selection format).
Like the ontology of ImageNet~\citep{deng2009imagenet}, no formalism
is perfect, but it enables automatic answer evaluation and
linking to a knowledge base.
In \qb{} though, the challenge though is not in framing an answer, it is in answering at the earliest possible moment.

The pyramidal construction of questions---combined with incrementality---makes \qb{} a more fair and granular comparison.
For example, the first sentence of Figure~\ref{fig:ex}---also known as the lead in---while obscure, uniquely identifies a single opera.
Questions that begin misleadingly are scorned and derided in online
discussions tournament as ``neg bait'';\footnote{ ``Negging'' refers
    to interrupting a question with a wrong answer; while wrong answers
    do happen, a response with a valid chain of reasoning should be
    accepted.  Only poorly written questions admit multiple viable
    answers.}
Thus, writers ensure that all clues are uniquely identifying even at
the start.

The entirety of questions are carefully crafted, not just the lead-in.
Middle clues reward knowledge but cannot be too easy: frequent clues
in questions or clues prominent in the subject's Wikipedia page are
considered ``stock'' and should be reserved for the end.
These same insights have been embraced by machine learning in the
guise of adversarial methods~\citep{Jia2017AdversarialEF} to
eschewing superficial pattern matching.
In contrast, the final giveaway clue should be direct and
well-known enough; someone with even a passing knowledge of \underline{The Magic Flute}
would be able to answer.

This is the product of a complicated and nuanced social dynamic in the
\qb{} community.
Top teams and novice teams often play on the same questions; questions are---in part---meant to teach~\citep{gall1970teaching} so are best when they
are fun and fair for all.
The pyramidal structure ensures that top teams use their deep
knowledge and quick thinking to buzz on the very first clues, but
novice teams are entertained and learning until they get to an
accessible clue.
Just about everyone answers all questions (it is considered a failure
of the question writer if the question ``goes dead'' without an
answer).

\qb{} is not just used to test knowledge; it also helps discover new
information and as a result diversifies questions (``oh, I did not know
the connection between the band the Monkees and correction
fluid!'').\footnote{ Bette Nesmith Graham, the mother of Monkees band
    member Michael Nesmith, invented correction fluid in 1956.  }
While most players will not recognize the first clue (otherwise the
question would not be pyramidal), it should be interesting and connect
to things the player would care about.
For example, in our \underline{Magic Flute} question, we learn that
the librettist appeared in the premiere, a neat bit of trivia that we
can tuck away once we learn the answer.

These norms have established \qb{} questions as a framework to both
test and educate human players.
Our thesis is that these same properties can also train and evaluate
machine question answering systems.
Next, we highlight the \abr{nlp} and \abr{ml} challenges in \qb{}.

\subsection{Quizbowl for Natural Language Processing Research}
\label{sec:challenges}

We return to Figure~\ref{fig:ex}, which exemplifies \abr{nlp} challenges common
to many \qb{} questions.
We already discussed (\textit{pyramidality}): each sentence uniquely
identifies the answer but each is easier than the last.
The most knowledgable answers earlier and ``wins'' the question.
But what makes the question difficult apart from
obscurity~\citep{2018arXiv180908291B}?
Answering questions early is significantly easier if machines can
resolve coreference~\citep{Ng2010SupervisedNP} and entity
linking~\citep{Shen2015EntityLW}.

First, the computer should recognize ``the librettist'' as
Schikaneder, \emph{whose name never appears in the question}.
%
This special case of entity linking to knowledge bases is sometimes called Wikification~\citep{Cheng2013RelationalIF,Roth2014WikificationAB}.
%
%
The computer must recognize that ``the librettist'' refers to a
specific person (mention detection), recognize that it is relevant to
the question, and then connect it a knowledge base (entity linking).

In addition to linking to entities \emph{outside} the question,
another challenge is connecting coreferences within a question.
The interplay between coreference and question answering is well known~\citep{stuckardt2003coreference}, but
\citet{Guha:Iyyer:Bouman:Boyd-Graber-2015} argue that \qb{} coreference is
particularly challenging: referring expressions are
longer and oblique, world knowledge is needed, and entities are named \emph{after}
other referring expressions.
Take the character Tamino (Figure~\ref{fig:ex}): while he is
eventually mentioned
by name, it is not until after he has been referred to obliquely (``a
man who claims to have killed a serpent'').
The character Papageno (portrayed by Schikaneder) is even worse; while
referred two twice (``character who asks for a glass of wine'', ``That
character''), Papageno is never mentioned by name.
To fully solve the question, a model may have to solve a difficult
coreference problem \textbf{and} link the reference to Papageno and
Schikaneder.

These inferences, like in the clue about ``the librettist'', are often
called \emph{higher-order reasoning} since they require creating and
combining inference rules to derive conclusions about multiple pieces
of information~\citep{Lin2001DiscoveryOI}.
Questions that require only a single lookup in a knowledge base or a
single \abr{ir} query are uninteresting for both humans and computers;
thus, they are shunned for \qb{} lead-in clues.
Indeed, the first sentences in \qb{} questions are the most difficult
clues for humans and computers because they often incorporate
surprising, quirky relationships that require skill and reasoning to
recognize and disentangle.
Interest in multi-hop question answering led to the creation WikiHop
through templates~\citep{Welbl2018ConstructingDF} and HotPotQA through
crowdsourcing~\citep{Yang2018HotpotQAAD}.
In contrast to these artificially or crowdsourced created datasets, \qb{} questions focus on
links that experts view as relevant and important.

Finally, even the final clue (called a ``giveaway'' because it's so
easy for humans) could pose issues for a computer.
Connecting ``enchanted woodwind instrument'' to \underline{The Magic
    Flute} requires solving wordplay.
While not all questions have all of these features, these features are
typical of \qb{} questions and showcase their richness.

Crowdsourced datasets like OpenBooksQA~\citep{mihaylov2018openbook}
and CommonSenseQA~\citep{talmor2019commonsense} have artifacts that
algorithms can game~\citep{geva2019annotator}: they find the right
answer for silly reasons.
For example, answering correctly with just a handful of words from a
\squad{} question~\citep{feng2018rawr}, none of a bAbI
question~\citep{DBLP:conf/emnlp/KaushikL18}, or the image in a
question about an image~\citep{Goyal2017MakingTV}.
Although the \qanta{} dataset and other ``naturally occurring'' data
likely do contain machine exploitable patterns, they do not face the
same quality issues since the author's motivation is intrinsic: to
write an entertaining and educational question as in \qb{}.


\subsection{Quizbowl for Machine Learning Research}
\label{sec:form-task}

While answering questions showcases the \abr{nlp} challenges, deciding
\emph{when} to answer showcases the \abr{ml} challenges related to decision theory~\citep{raiffa1968decision}.
As in games like Poker~\citep{brown2019poker}, \qb{} players have incomplete information: they do not know when their opponent will answer, do not know what clues will be revealed next, or if they will know the next clues.
%
%
In our buzzer model, the \qa{} model output is but one piece of information used to make the decision---under uncertainty---of when to buzz in.
Since a decision must be made at every time step (word), we call this an incremental classification task.

We formalize the incremental classification task as a Markov Decision
Process~\citep[\abr{mdp}]{DBLP:conf/icml/ZubekD02}.
The actions in this \abr{mdp} correspond to what a player can do in a
real game: click the buzzer and provide their current best answer or
wait (one more word) for more information.
The non-terminal states in the state space are parameterized by the
text of the question revealed up to the current time step, the
player's current best guess, and which player (if any) has already
buzzed incorrectly.
Rewards are only given at terminal states and transitions to those
states are determined by which player correctly answered first.
Additionally, we treat the opponent as a component of the environment
as opposed to another agent in the game.\footnote{ This is not
    precisely true in our live exhibition matches; although we treat the
    opponent as part of the environment, our human opponents do not and
    usually adapt to how our system plays.  For instance, it initially
    had difficulty with pop culture questions. } This task---the buzzing
task---has connections to work in model confidence calibration offline~\citep{yu2011calibration,nguyen2015calibration} as well as online~\citep{kuleshov2017estimating}, cost-sensitive
learning~\citep{elkan2001cost}, acquisition of features with a budget~\citep{lizotte2012budget}, and incremental classification~\citep{melville2005feature}.

For humans, effective \qb{} play involves maintaining a correctness
estimate of their best answer, weighing the cost and benefits of
answering now versus waiting, and making buzzing decisions from this
information.
Naively, one might assume that model calibration is as simple as
examining the probability output by the (neural) \qa{} system, but
neural models are often especially poorly
calibrated~\citep{Guo:Pleiss:Sun:Weinberger-2017} and calibrations
often fail to generalize to out of domain test
data~\citep{kamath2020selective}.
Since \qb{} training data spans many years, models must also contend
with domain shift~\citep{ovadia2019trust}.
Model calibration is naturally related to deciding when to buzz---also
known as answer triggering in \qa{} and information
retrieval~\citep{Voorhees2001OverviewOT,Yang2015WikiQAAC}.

Unlike standard answer triggering though, in \qb{} the expected costs
and benefits are continually changing.
Specifically, there are costs for obtaining new information (seeing
more words) and costs for misclassifications (guessing incorrectly or
waiting too long).
This parallels the setting where doctors iteratively conduct medical
tests until they are confident in a patient's
diagnosis~\citep{DBLP:conf/icml/ZubekD02,Chai2004TestcostSN}.

Although this can be framed as reinforcement learning, we instead
frame buzzing in Section~\ref{sec:buzzing} as incremental
classification as in \citet{trapeznikov2013sequential}.
In this framing, a binary classifier at each time step determines when
to stop obtaining new information and render the decision of the
underlying (\qa{}) model.
As \citet{trapeznikov2013sequential} note, evaluation in this scenario
is conceptually simple: compare the costs incurred to benefits gained.

\paragraph{Evaluation}

We evaluate the performance of our systems through a combination of standalone
comparisons (Section~\ref{sec:guess-eval}) and simulated \qb{} matches
(Section~\ref{sec:buzz-eval}).
For standalone evaluation we incrementally feed
systems new words and record their responses.
We then calculate accuracy for each position in the question (e.g.,
after the first sentence, halfway through the question, and at the
end).
While standalone evaluations are useful for developing systems, the best way
to compare systems and humans is with evaluations that mimic \qb{}
tournaments.

\ignore{
    To play simulated games, we assume that a match consists of a sequence
    of questions called a packet.
    Each question is revealed incrementally to the players (machine or
    human) until one decides to buzz in with an
    answer.\footnote{Ties---which are infrequent---are broken randomly.}
    If the answer is correct the player gains ten points.
    Otherwise, they lose five points and their opponent has an opportunity
    to answer.\footnote{The opponent can answer based on the full
        question.  I.e., they do not have to answer ``early'' once the first
        player gets it wrong.
    }
    Once all the questions in the packet have been played, the player with
    the most points wins.
    In our full evaluations, we have systems play matches against each
    other round-robin style, play simulated matches against humans using
    the gameplay dataset, and play live exhibition matches against teams
    of accomplished trivia players.
}

A recurring theme is our mutually beneficial collaboration with the
\qb{} community: host outreach exhibitions (Section~\ref{sec:live}),
annotate data, play with and against our
systems (Section~\ref{sec:augment}), and collect the \qanta{} dataset.
This community created this rigorous format for question answering
over decades and continues to help understand and measure the question
answering abilities of machines.

%% file: 2021_jmlr_qanta/sections/30-dataset.tex
\section{QANTA Dataset}
\label{sec:datasets}

This section describes the \qanta{} dataset from the \qb{} community
(Section~\ref{sec:sources}).
The over 100,000 human-authored, English questions from \qb{} trivia
tournaments (Section~\ref{sec:qanta-questions}) allows systems to learn what to answer.
More uniquely, 3.9 million filtered records of humans playing \qb{} online (Section~\ref{sec:protobowl-data})
allows systems learn when to ``buzz in'' against opponents (Section~\ref{sec:tsm}).


\subsection{Dataset Sources}
\label{sec:sources}

The \qb{} community maintains and curates several public databases of
questions spanning 1997 to today.\footnote{
  Questions in were obtained (with permission) from \url{http://quizdb.org} and \url{http://protobowl.com}.
}
On average, 10,000 questions are written every year.
%
Our dataset has 119,247 questions with over 650 thousand sentences and 11.4
million tokens.

\begin{figure}
  \centering
  \begin{subfigure}{\textwidth}
    \centering
    \includegraphics[width=0.5\linewidth]{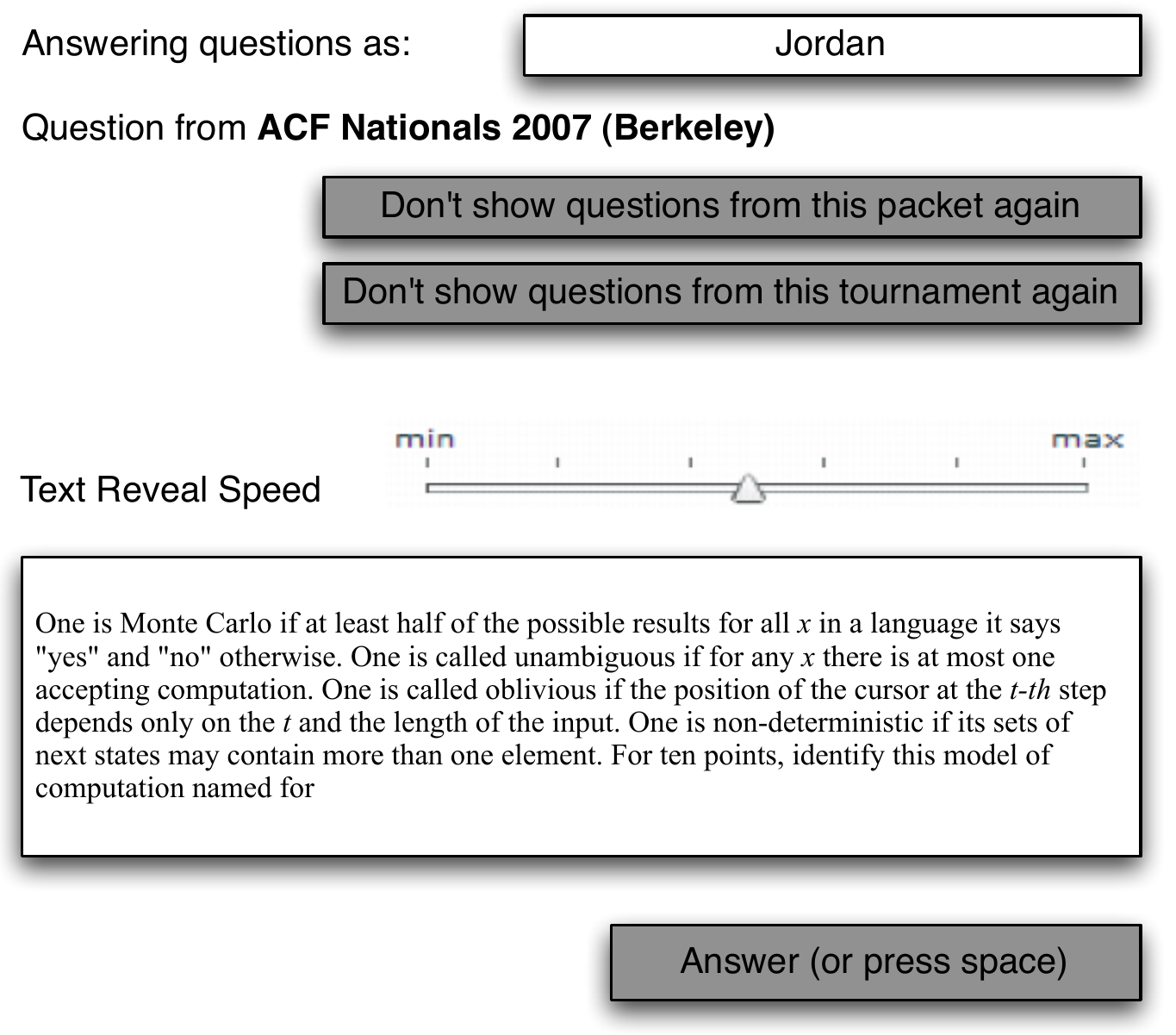}
    \caption{Our 2012 interface was the first way to play \qb{} online.}
    \label{fig:int-orig}
  \end{subfigure}\hfill
  \begin{subfigure}{\textwidth}
    \centering
    \includegraphics[width=1.0\linewidth]{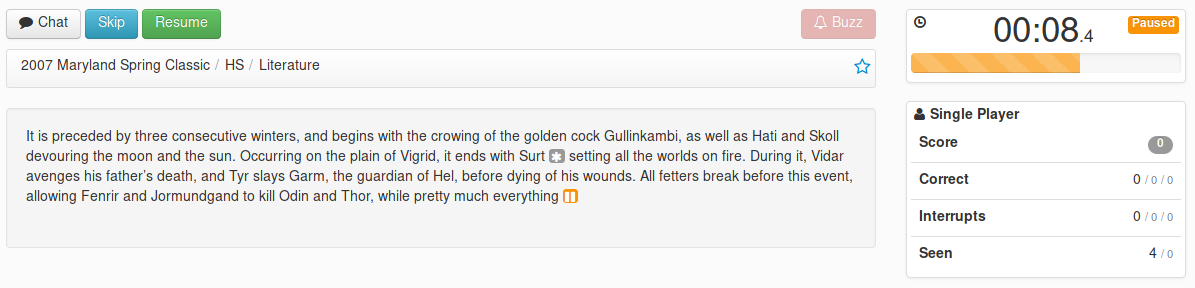}
    \caption{
      The \qb{} interface for collecting most of our gameplay records.
      It improved over our own through features like real-time competitive play and chatrooms.}
    \label{fig:int-proto}
  \end{subfigure}
  \caption{ Our interface and a popular modern interface for playing
    \qb{} online.  Both interfaces reveal questions word-by-word until
    a player interrupts the system and makes a guess.  }
  \label{fig:interface}
\end{figure}

To help players practice and to build a dataset showing how humans
play, we built the first website for playing \qb{} online
(Figure~\ref{fig:int-orig}).
After initial popularity, we shut down the site; however, enterprising
members of the \qb{} community
resurrected and improved the application.
89,477 players used the successor (Figure~\ref{fig:int-proto}) and
have practiced 5.1 million times on 131,075 unique questions.
A filtered\footnote{
  We include only a player's first play on a question and exclude players with less than twenty questions.} subset of 3.9 million player
records forms the second component of our dataset, which we call
\textbf{gameplay data}.

\subsection{QANTA Questions}
\label{sec:qanta-questions}

%
Table~\ref{table:datasets} compares \abr{qa} datasets written by humans.
Because often each \qb{} sentence has enough information for players
to answer, each \qanta{} instance can be broken into four to six pseudo
sentence-answer pairs.
Although our dataset does not have the most questions, it is
significantly larger in the number of sentences and tokens.

\begin{table}
  \centering
  \small
  \input{2021_jmlr_qanta/auto_fig/dataset_comparison.tex}
  \caption{The \qanta{}~dataset is larger than most question answering
    datasets in \abr{qa} pairs (120K).
    However, for most \qb{} instances each sentence in a question can be
    considered a \abr{qa} pair so the true size of the dataset is
    closer to 650K QA pairs.
    In Section~\ref{sec:guessing} using sentence level \abr{qa} pairs
    for training greatly improves model accuracy.
    The \qanta{}~dataset has more tokens than all other
    \abr{qa} datasets.
    Statistics for \qanta{} 2012 and 2013 only include publicly available
    data.
  }
  \label{table:datasets}
\end{table}

\begin{figure*}[t]
  \centering \includegraphics[width=\textwidth]{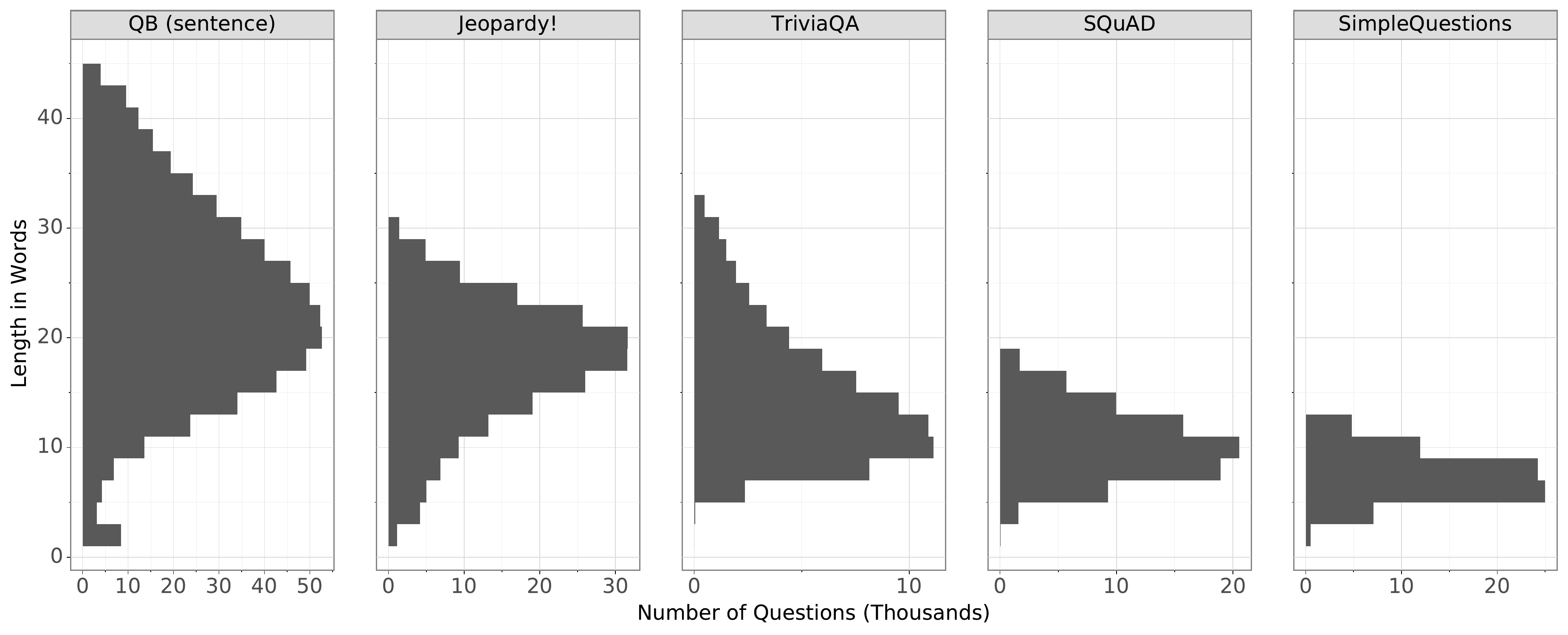}
  \captionof{figure}{
    Size of question answering datasets.
    Questions in the \qanta{} dataset have longer sentences than any other dataset.
    The instances from SimpleQuestions, SQuAD, and \triviaqa{} are comparatively short which makes it less likely that they are as diverse of \qb{} or Jeopardy!.
    For each dataset we compare the lengths of questions rather than paired context paragraphs; to avoid the histogram being overly skewed we remove the top 5\% of examples by length from each dataset.
  }
  \label{fig:length-dist}
\end{figure*}

In addition to \qanta{} having more sentences, questions are longer
(Figure~\ref{fig:length-dist}), especially compared to crowd-sourced
datasets.
As a side effect of both being longer and not crowdsourced, \qb{}
sentences are syntactically complex and topically diverse (Figure~\ref{fig:category-dist}).
%

\subsubsection{Dataset Diversity}
\label{sec:topical-div}

Creating diverse datasets is a shared goal between researchers
developing \abr{nlp} resources and organizers of \qb{} tournaments.
\qb{} questions are syntactically diverse with dense
coreference~\citep{Guha:Iyyer:Bouman:Boyd-Graber-2015} and cover a
wide range of topics.
Diversity takes the form of questions that reflect the topical,
temporal, and geographical breadth of a classical liberal education.
For example, the Academic Competition Federation mandates that
literature cover American, British, European, and world
literature~\citep{acf-distro}.
Moreover, authors must ``vary questions across time periods''---with no
more than one post 1990 literature---and questions must ``span a
variety of answers such as authors, novels, poems, criticism, essays,
etc.''
There are similarly detailed proscriptions for the rest of the
distribution.

\begin{figure*}
  \centering
  \includegraphics[width=.7\textwidth]{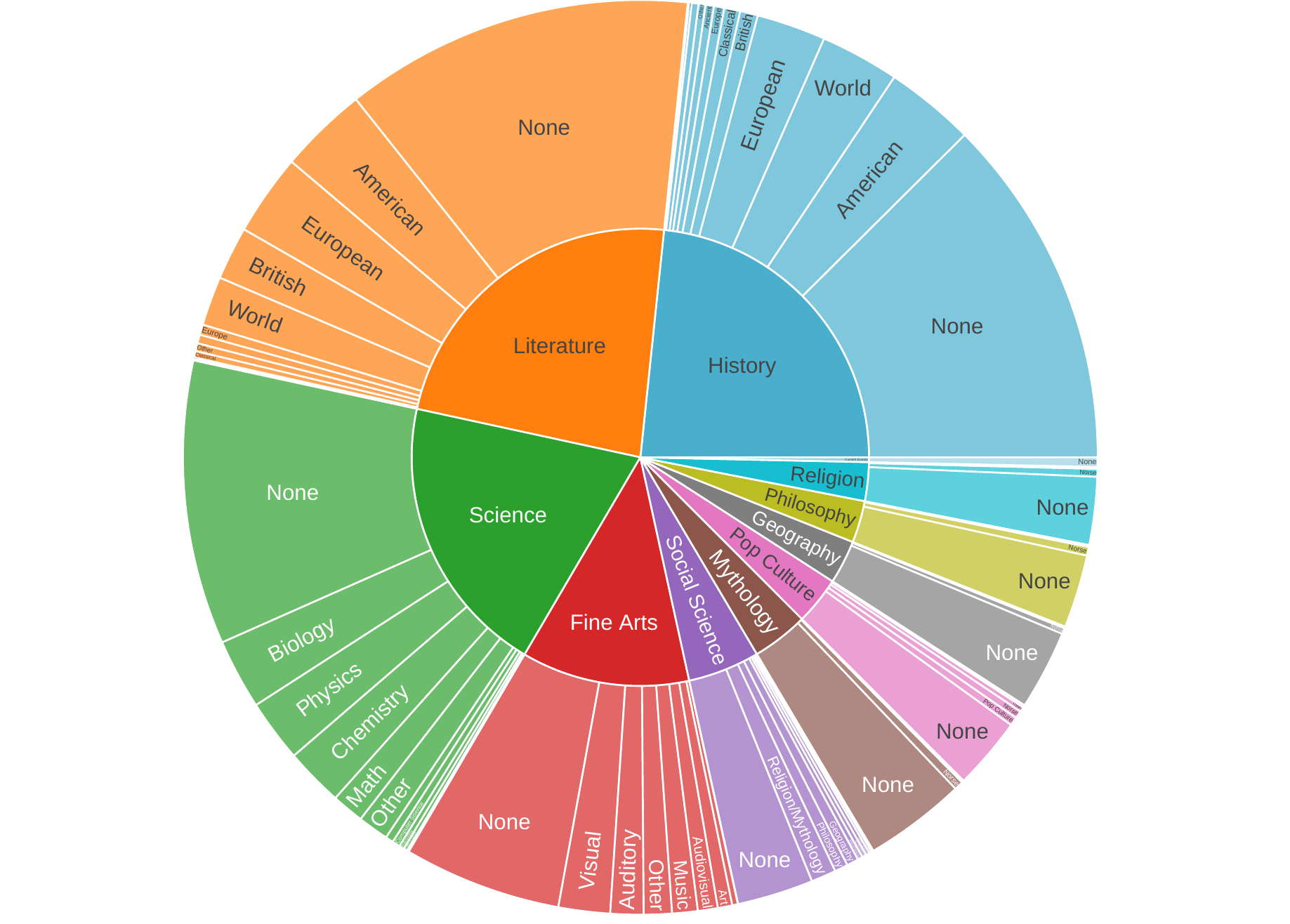}
  \caption{
    Questions in \qb{} cover most if not all academic topics
    taught in school such as history, literature, science, the fine arts, and social sciences.
    Even within a single category, questions cover a range of topics.
    Topically, the dataset is biased towards American and European topics in literature and history.
  }
  \label{fig:category-dist}
\end{figure*}

Figure~\ref{fig:category-dist} shows the category and sub-category
distribution over areas such as history, literature, science, and fine
arts.
Taken together, \qb{} is a topically diverse dataset across broad
categories and finer-grained sub-categories.
This diversity contrasts with a sample of $150$ questions from NaturalQuestions~\citep{nq19}\footnote{
  The authors annotated $150$ questions from the development set using the same categories as \qb{}.
} which indicates that questions are predominantly about Pop Culture ($40\%$), History ($19\%$), and Science ($15\%$); see Appendix~\ref{apx:nq} for complete results.
This emphasizes that to do well, players and systems need to have both
breadth and depth of knowledge.

\subsubsection{Answer Diversity}
\label{sec:answer-div}

\begin{figure*}
  \centering
  \includegraphics[width=.7\textwidth]{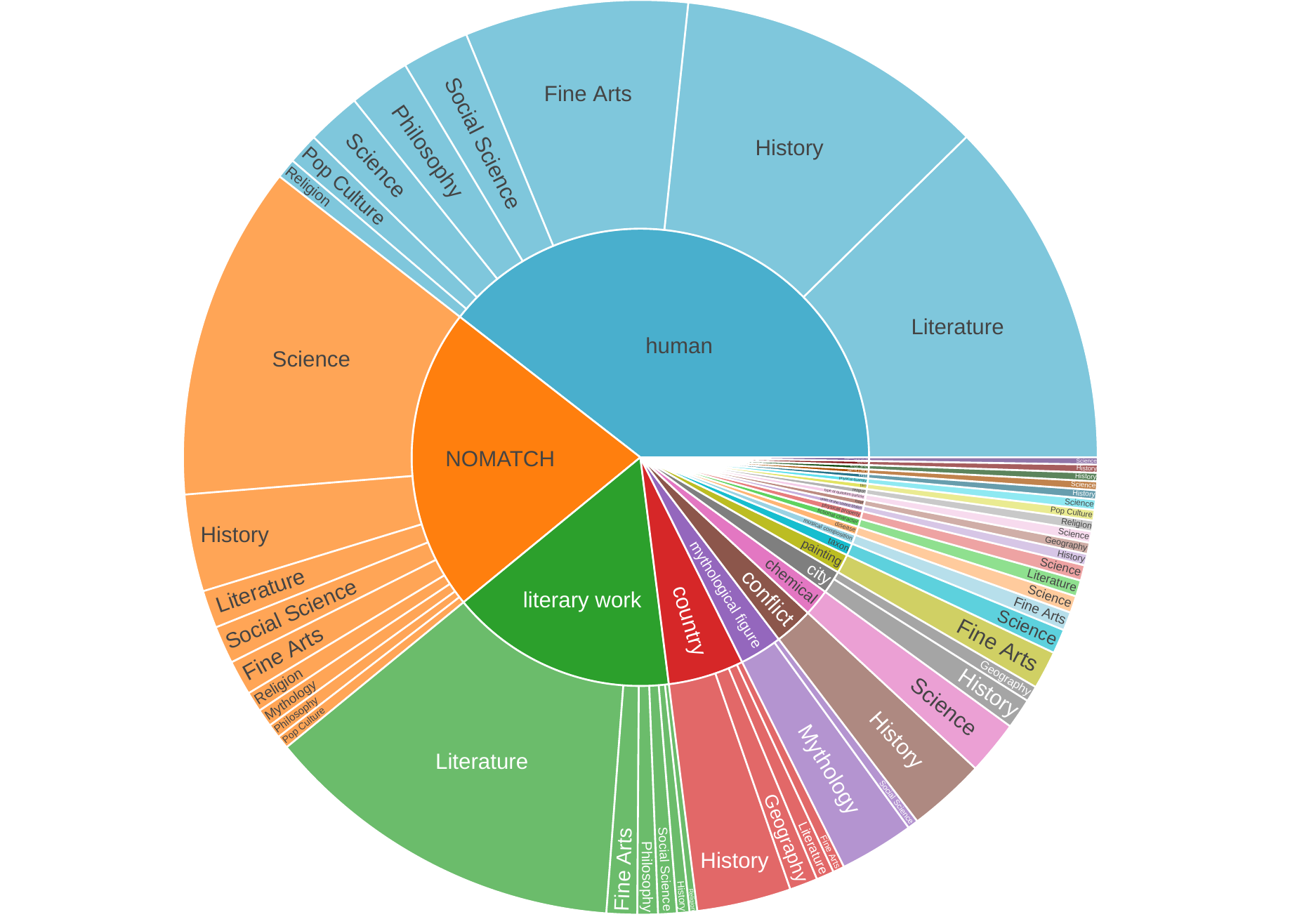}
  \captionof{figure}{
    Distribution of wikidata.org answer types (``instance of'' relation) further broken down by category.
    Most answers have matching types and reference a person, literary work, or geographic entity.
    Among these types, there is a good balance of answers spread across literature, history, fine arts, and science.
    Answer types with only one category are largely self-explanatory (e.g., mythological answers types to the mythology category).
    The special category ``NOMATCH'' are answers without a matched type and similar types are merged into larger categories.
  }
  \label{fig:answer-type-dist}
\end{figure*}

\qb{} questions are also diverse in the kinds of entities that appear
as answers (25K entities in the training data).
A dataset which is topically diverse, but only asks about people is
not ideal.
Using the Wikidata knowledge graph we obtain the type of each answer
and plot frequencies in
Figure~\ref{fig:answer-type-dist}.
Most questions ask about people (human), but with a broad diversity
among other types.

These two breakdowns show that \qb{} is topically and answer-wise
diverse.
To \qb{} aficionados this is unsurprising; the primary educational
goal of \qb{} is to encourage students to improve their mastery over
wide ranges of knowledge.
We now turn to details about the gameplay dataset.

\subsection{Gameplay Records}
\label{sec:protobowl-data}

Like the $2002$ \abr{trec} \abr{qa} track~\citep{Voorhees2004-im},
\squad{} 2.0~\citep{Rajpurkar2018KnowWY}, and \nq{}~\citep{nq19},
deciding when \emph{not} to answer is crucial to playing \qb{}.
Unlike these tasks though, deciding when to answer is not just model
calibration or triggering, but also should reflect the opponent's
behavior~\citep{billings1998poker}.
To address this, we use gameplay data
(Table~\ref{table:protobowl_entry}) which contain records of quizbowlers playing questions from prior
tournaments: words in each question were revealed one-by-one until the
player guessed the question's answer.
We use these records as (1) training data so that models can learn
to imitate an oracle buzzing
policy~\citep{coates2008imitation,ross2010reduction,ross2011noregret}
and (2) as human baselines for offline evaluations
(Section~\ref{sec:eval}).

\begin{table}
  \centering
  \begin{tabular}{lp{0.6\textwidth}}
    \toprule
    Date          & Thu Oct 29 2015 08:55:37 GMT-0400 (EDT)                \\
    UID           & 9e7f7dde8fdac32b18ed3a09d058fe85d1798fe7               \\
    QID           & 5476992dea23cca90550b622                               \\
    Position      & 47                                                     \\
    Guess         & atlanta                                                \\
    Result        & True                                                   \\
    Question text & This Arcadian wounded a creature sent to punish Oeneus
    for improperly worshipping Artemis and killed the centaurs Rhaecus and
    Hylaeus\ldots                                                          \\
    \bottomrule
  \end{tabular}
  \caption{An entry from the gameplay dataset where the player correctly
    guesses ``Atlanta'' at word 47. The entry \abr{qid} matches with the
    \abr{proto\_id} field in the question dataset where additional
    information is stored such as the source tournament and year.}
  \label{table:protobowl_entry}
\end{table}

Like \citet{mandel2014offline}, gameplay records both
simulate humans for training and evaluating policies.
To simulate play against a human, we see which agent---human or
machine---first switches from the wait action to the buzz action.
For example, in Table~\ref{table:protobowl_entry} the user correctly
guessed ``Atlanta'' at word forty-seven.
If an agent played against this player they would need to answer
correctly before word forty-seven to win.
In all but one outcome, replaying the human record exactly recreates a
live face-off.
When a machine incorrectly buzzes first we lack what the human would
ultimately guess, so we assume their guess would have been correct
since skilled players almost always answer correctly by the end of the
question.
During training, these data help agents learn optimal buzzing policies based on their own uncertainty, the questions, and their opponents' history~\citep{he2016opponent}.\footnote{
In this article, we significantly expand the number of player-question records.
We also make the setting significantly harder by not restricting questions to only the most frequently asked about answers (1K versus 24K).
Finally, we create a new evaluation procedure (Section~\ref{sec:curve_score}) that better estimates how models fare in the real-world versus human players.
The first version of the gameplay dataset and models was introduced in:
\\ He He, Jordan Boyd-Graber, and Hal
Daum\'{e} III.  {\bf
    \href{http://umiacs.umd.edu/~jbg//docs/2015_acl_dan.pdf}{Opponent Modeling in Deep Reinforcement Learning}}. \emph{International Conference on Machine Learning}, 2016.
}

With this data, we compute how models would fare against human players
individually, players partitioned by skill, and in expectation
(Section~\ref{sec:curve_score}).
In contrast to this strategy, crowdsourced tasks (e.g., \squad{})
often use the accuracy of a single annotator to represent human
performance, but this is problematic as it collapses the distribution
of human ability to a single crowd-worker and does not accurately
reflect a task's upper bound compared to multiple
annotation~\citep{nangia2019muppet,nq19}.
In the gameplay data, we have ample data with which to robustly
estimate average and sub-group human skill; for example, 90,611 of the
131,075 questions have been played at least five times.
This wealth of gameplay data is one aspect of \qb{}'s strength for
comparing humans and machines.

An additional aspect unique to trivia games is that participants are
intrinsically motivated experts.
Compensation---i.e., extrinsic motivation---in crowdsourcing is
notoriously difficult.
If they feel underpaid, workers do not give their best
effort~\citep{gneezy2000pay}, and increasing pay does not always
translate to quality~\citep{mason2009incentives}.
In light of this, \citet{mason2009incentives} recommend intrinsic
motivation, a proven motivator for annotating
images~\citep{ahn2004label} and protein
folding~\citep{cooper2010protein}.
Second, although multiple non-expert annotations can approach gold
standard annotation, experts are better participants when
available~\citep{snow2008cheap}.
Thus, other tasks may understate human performance with
crowdworkers lacking proper incentives or skills.
%

\begin{figure}[t]
  \begin{minipage}{.5\linewidth}
    \centering
    \includegraphics[width=\linewidth]{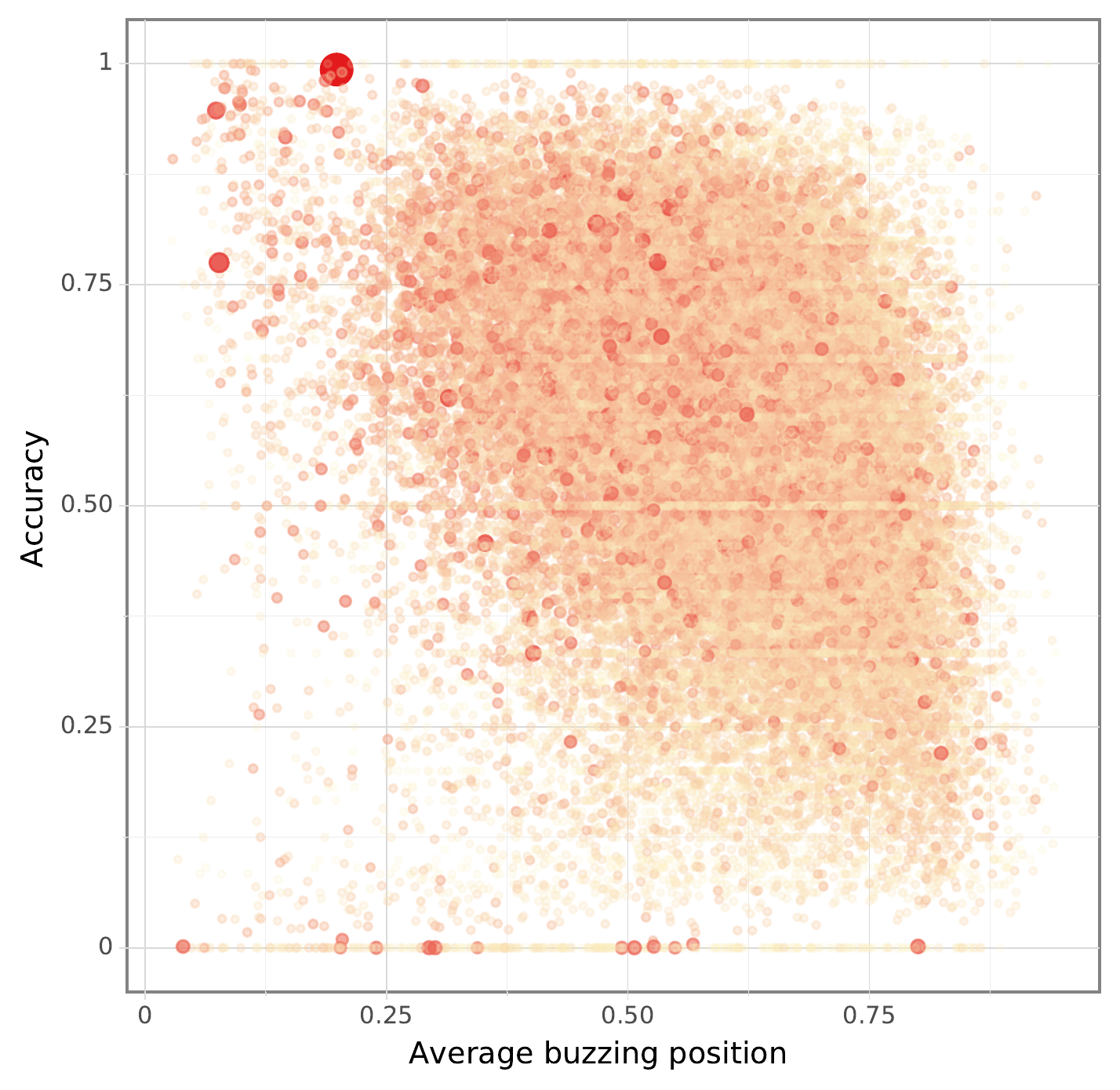}
  \end{minipage}\hspace{0.05\linewidth}%
  \begin{minipage}{.45\linewidth}
    \centering
    \includegraphics[width=\linewidth]{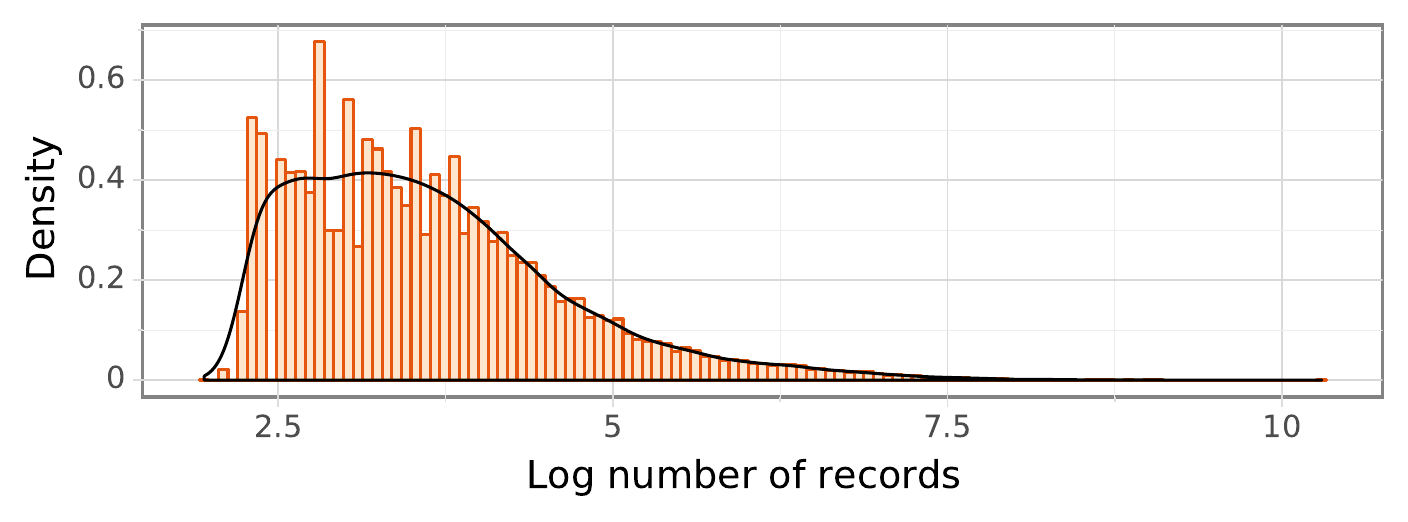}
    \includegraphics[width=\linewidth]{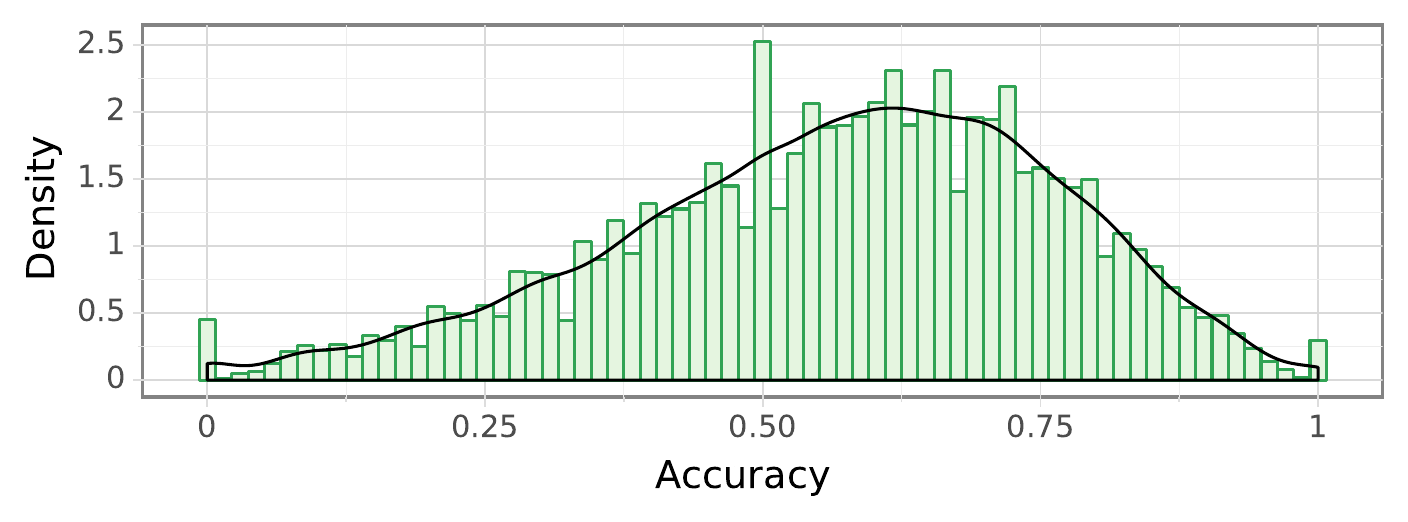}
    \includegraphics[width=\linewidth]{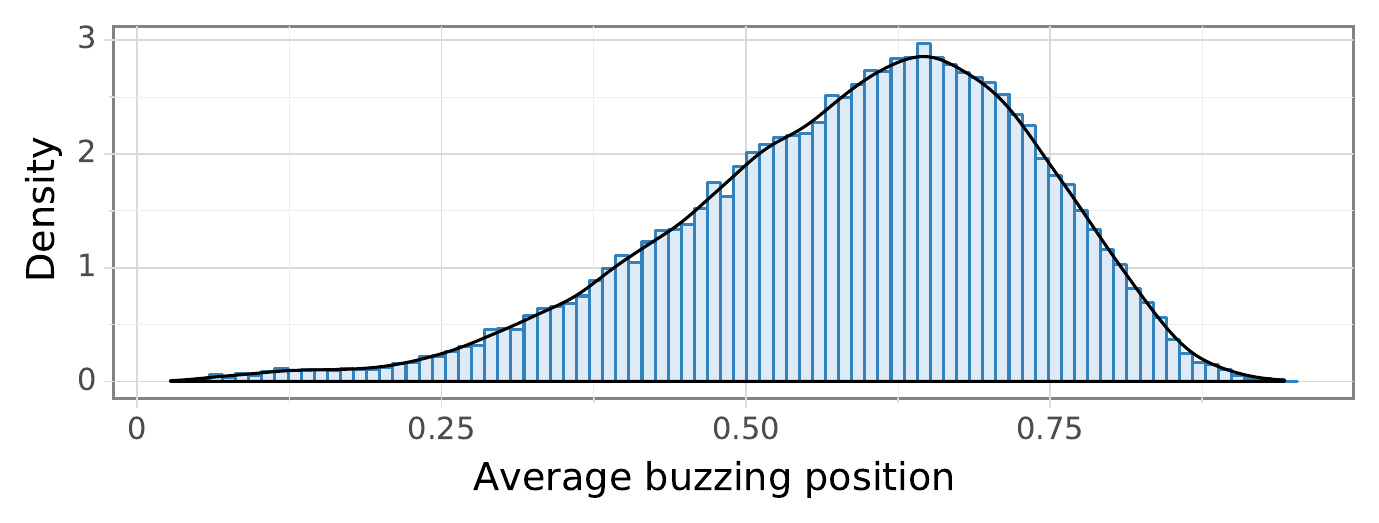}
  \end{minipage}
  \caption{Left: each protobowl user is represented by a dot,
    positioned by average accuracy and buzzing position; size and
    color indicate the number of questions answered by each user.
    Right: distribution of number of questions answered, accuracy,
    and buzzing position of all users.  An average player buzzes
    with 65\% of the question shown, and achieves about 60\%
    accuracy.}
  \label{fig:protobowl_users}
\end{figure}

Good quizbowlers are both accurate and quick.
To measure skill, we compute and plot in
Figure~\ref{fig:protobowl_users} the joint distribution of average
player accuracy and buzzing position (percent of the question
revealed).
The ideal player would have a low average buzzing position (early
guesser) and high accuracy; thus, the best players reside in the upper
left region.
On average, players buzzes with 65\% of the question shown with 60\%
accuracy (Figure~\ref{fig:protobowl_users}).
Although there are other factoid \qa{} and---more
specifically---trivia datasets, \qb{} is the first and only dataset
with a large dataset of gameplay records which allows us to train
models and run offline benchmarks.

\subsection{Preprocessing}
\label{sec:q-proc}

Before moving to model development, we describe necessary
preprocessing to questions eliminate answer ambiguity, pair questions
to gameplay data, and creating dataset folds that enable independent
yet coordinated training of distinct guessing and buzzing models.
Preprocessing is covered in significantly more detail in
Appendix~\ref{apx:preprocess}.

\paragraph{Matching QB Answers to Wikipedia Pages}
Throughout this work we frame \qb{} as a classification task over the set of Wikipedia page entities (Section~\ref{sec:craft}), which necessarily requires pairing answers to distinct pages if on exists.
We pair questions and their answers to Wikipedia pages in two steps: parsing potential answers from moderator instructions and matching to Wikipedia entities.\footnote{
  We preprocess the English Wikipedia \wikidumpdate{} dump with \url{https://github.com/attardi/wikiextractor}.
}
In \qb{}, the ``answers'' are in actuality instructions to the moderator that may provide additional detail on what answers are acceptable.
For example, answer strings like ``Second Vatican Council [or Vatican
    II]'' indicate to accept either surface form of the same concept.
Fortunately, the vast majority of these ``answer instructions'' are
automatically parsable due to their semi-regular structure.
The second step---described further in
Appendix~\ref{apx:matching}---matches parsed answers to pages through
a combination of strict textual matching, expert-curated matching
rules (e.g., only match ``camp'' to \underline{Camp\_(style)} if
``style'' or ``kitsch'' are mentioned), and expert annotated pairings
between questions and pages.\footnote{ Primarily, the authors of this
  article annotated the answer to page pairings.  }
In total, we paired \nquestions{} out of \ntotalquestions{} with
Wikipedia titles (examples in Appendix~\ref{apx:matching}).

\paragraph{Dataset Folds}

The goal of the folds in the \qanta{} dataset is to standardize the
training and evaluation of models for the guessing and buzzing
sub-tasks.
Towards this goal, we sub-divide the \qanta{} dataset by sub-task and
standard machine learning folds (e.g., training, development, and
test).
We create the standard machine learning folds by partitioning the data
according to tournament type and year.
To increase the quality of evaluation questions, we only include questions from championship level tournaments in the development and test folds.\footnote{We use questions from \textsc{acf} Regionals, \textsc{acf} Nationals,
  \textsc{acf} Fall, \textsc{pace nsc}, and \textsc{nasat} from 2015 onward for
  development and test sets.}
To derive the final folds, we temporally divide the data~\citep{arlot2010cross} so that only (championship) questions from 2015 and onward are used in evaluation folds.

The subdivision by task simultaneously addresses the issue that some
questions lack gameplay data (thus are not helpful for buzzer
training) and partitioning the data so that the buzzer calibrates
against questions unseen during training (details in
Appendix~\ref{apx:f-assign}).
Table~\ref{table:dataset-folds} shows the size of each sub-fold; unassigned questions correspond to those where the answer to page matching process failed.
Finally, hundreds of new \qb{} questions are created every year which provides an opportunity for continually adding new training questions and replacing outdated test questions.
Ultimately, this may help temper overconfidence in the generalization of models~\citep{Patel2008InvestigatingSM} since we expect there to be covariate shift, prior probability shift, and domain shift in the data~\citep{quionero2009shift} as questions evolve to reflect modern events.

\begin{table}
  \centering
  \begin{tabular}{ l r }
    \toprule
    Fold          & Number of Questions \\
    \midrule
    train + guess & $96,221$            \\
    train + buzz  & $16,706$            \\
    dev + guess   & $1,055$             \\
    dev + buzz    & $1,161$             \\
    test + guess  & $2,151$             \\
    test + buzz   & $1,953$             \\
    unassigned    & $13,602$            \\
    \midrule
    All           & $132,849$           \\
    \bottomrule
  \end{tabular}
  \caption{We assign each question in our dataset to either the train,
    development, or test fold. Questions in the development and test folds come
    from national championship tournaments which typically have the highest quality
    questions. The development and test folds are temporally separated from the
    train and development folds to avoid leakage. Questions in each fold are
    assigned a ``guess'' or ``buzz'' association depending on if they have gameplay
    data. Unassigned refers to questions for which we could not map their answer
    strings to Wikipedia titles or there did not exist an appropriate page to match
    to.}
  \label{table:dataset-folds}
\end{table}

The \qanta{} datasets, a copy of the Wikipedia data used, intermediate
artifacts, and other related datasets are available at
\url{http://datasets.qanta.org}.

%% file: 2021_jmlr_qanta/auto_fig/dataset_comparison.tex
\begin{tabular}{ l r r} \toprule Dataset & QA Pairs (Sentences / Questions) &
Tokens\\ \midrule \simplequestions{}~\citep{DBLP:journals/corr/BordesUCW15} &
100K & .614M\\ \triviaqa{}~\citep{JoshiTriviaQA2017} & 95K & 1.21M\\
\squad{}~1.0 \citep{DBLP:conf/emnlp/RajpurkarZLL16} & 100K & .988M\\
\searchqa{}~\citep{Dunn2017SearchQAAN} & 216K & 4.08M\\
NaturalQuestions~\citep{nq19} & 315K & 2.95M\\ \midrule  \qanta{}~2012 \citep{Boyd-Graber:Satinoff:He:Daume-III-2012} & 47.8K / 7.95K  & 1.07M\\  \qanta{}~2014 \citep{Iyyer:Boyd-Graber:Claudino:Socher:Daume-III-2014} & 162K / 30.7K  & 4.01M\\  \qanta{}~2018 \textbf{(This Work)}&
\textbf{650K / 120K} & \textbf{11.4M}\\ \bottomrule \end{tabular}

%% file: 2021_jmlr_qanta/sections/40-system.tex
\section{Deciding When and What to Answer}
\label{sec:tsm}
\input{2021_jmlr_qanta/sections/45-tsm-figure.tex}

One could imagine many machine learning models for playing \qb{}: an
end-to-end reinforcement learning model or a heavily pipelined model
that determines category, answer type, answer, and decides when to
buzz.
Without making any value judgment on the \emph{right} answer, our
approach divides the task into two subsystems: {\bf guessing} and {\bf
    buzzing} (Figure~\ref{fig:guesser-buzzer-system}).
This approach mirrors \abr{ibm} Watson's\footnote{In Watson, the
  second system also determines wagers on Daily Doubles, wagers on
  Final Jeopardy, and chooses the next question (e.g., history for
  \$500)} two model
design~\citep{Ferrucci2010BuildingWA,Tesauro2013AnalysisOW}.
The first model answers questions, and the second decides when to
buzz.
Dividing a larger task into sub-tasks is common throughout machine
learning, particularly when the second is making a prediction based on
the first's prediction.
For example, this design pattern is used in object detection
\citep[generate bounding box candidates then classify
  them]{girshick2014rcnn}, entity linking~\citep[generate candidate
  mentions and then disambiguate them to knowledge base
  entries]{ling2015design}, and confidence estimation for automatic
speech recognition systems~\citep{kalgaonkar2015asr}.
In our factorization, guessing is based solely on question text.
At each time step (word), the guessing model outputs its best guess,
and the buzzing model determines whether to buzz or wait based on the
guesser's confidence and features derived from the game state.
This factorization cleanly reduces the guesser to question answering
while framing the buzzer as a cost-sensitive confidence calibrator.

This division of modeling labor makes it significantly easier to train
the buzzer as a learned calibrator of the guesser's softmax classifier
predictions.
This is crucial since the probabilities in neural softmax classifiers
are unreliable~\citep{Guo:Pleiss:Sun:Weinberger-2017}.
Like how we train a calibration model (buzzer) over a
classifier (guesser), \citet{corbiere2019conf} train a calibration
model on top of a image classification model which is a more effective
approach in high dimensional spaces compared to nearest-neighbor based
confidence measures~\citep{jiang2018trust}.
However, not all buzzing errors are equal in severity; thus, part of the buzzer's challenge is in incorporating cost-sensitive classification.
By partioning model responsibilities into separate guessing and
buzzing models, we can mitigate the calibration-based drawbacks of
neural softmax classifiers while naturally using gameplay data for
cost-sensitive decision-making.

Machines playing \qb{} by guessing and buzzing semi-independently is
also convenient from an engineering perspective: it simplifies model
training and is easier to debug.
More importantly, it allows us and subsequent researchers to focus on
a sub-task of their choosing or the task as a whole.
If you are interested in only question answering, focus on the
guesser.
If you are interested in multiagent cooperation or confidence
estimation, focus on the buzzer.
Following the discussion of our guessing (Section~\ref{sec:guessing})
and buzzing (Section~\ref{sec:buzzing}) systems we describe our
evaluations and results in Section~\ref{sec:guess-eval}.
Section~\ref{sec:live} summarizes the outcomes of our live, in-person,
exhibition matches against some of the best trivia players in the
world.

%% file: 2021_jmlr_qanta/sections/45-tsm-figure.tex
\begin{figure*}
  \centering
  \includegraphics[width=\textwidth]{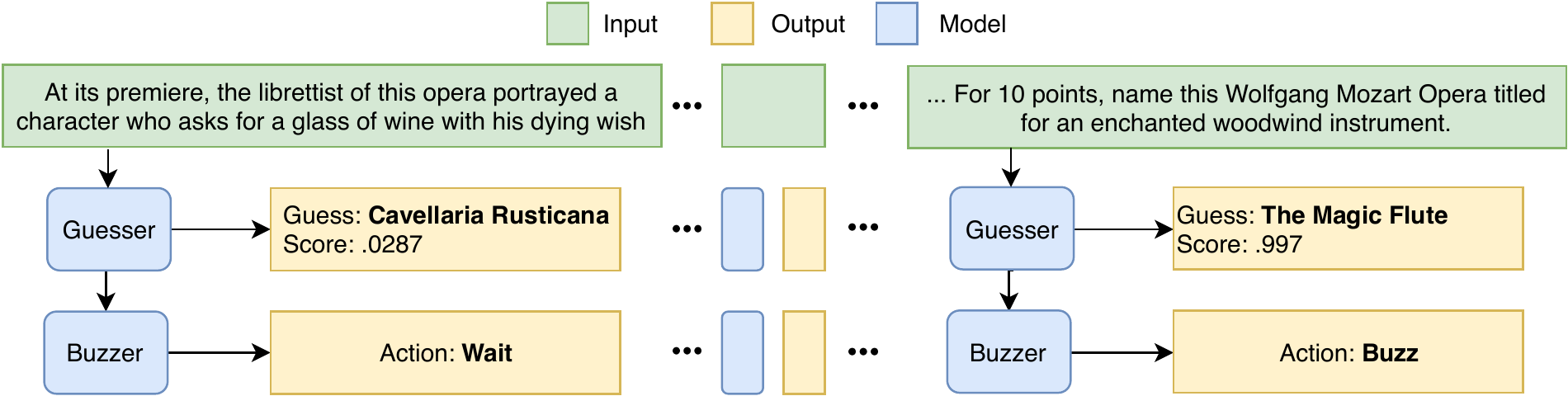}
  \caption{The \qanta{} framework for playing Quiz Bowl with semi-independent
    guesser and buzzer models. After each word in the input is revealed the
    guesser model outputs its best guesses. The buzzer uses these in combination
    with positional and gameplay features to decide whether to take the buzz or
    wait action. The guesser is trained as a question answering system that
    provides guesses given the input text seen so far.  Buzzers take on dual roles as calibrators of the guesser confidence scores and cost-sensitive decision classifiers by using the
    guesser's score, positional features, and human gameplay data.}
  \label{fig:guesser-buzzer-system}
\end{figure*}

%% file: 2021_jmlr_qanta/sections/50-guessing.tex
\section{Guessing QB Answers}
\label{sec:guessing}

Guessing answers to questions is a factoid question answering task and
the first step towards our models playing \qb{}
(Figure~\ref{fig:guesser-buzzer-system}).
We frame the question answering sub-task in \qb{} as high dimensional
multi-class classification over Wikipedia entities (i.e., answers are
entities defined by distinct Wikipedia pages).
This section describes three families of question answering models:
information retrieval models (Section~\ref{sec:ir}), linear models
(Section~\ref{sec:linear}), and neural models
(Section~\ref{sec:neural}).
Despite distinct differences, these approaches share a common
structure: create a vector representation $\bm{x}$ of the input
question, create a vector representation for each candidate answer
$\bm{a}_i$, and then return the answer $A_i$ corresponding to
$\argmax_i{f(\bm{x}, \bm{a}_i)}$ where $f$ is some similarity
function.\footnote{For brevity and clarity, we omit bias terms.}

\subsection{Explicit Pattern Matching with Information Retrieval}
\label{sec:ir}

The first model family we discuss are traditional information
retrieval (\textsc{ir}) models based on the vector space
model~\citep{salton1975vector}.
Vector space models are particularly effective when term overlap is a
useful signal---as in factoid \qb{}~\citep{lewis2020overlap}.
For example, although early clues avoid keyword usage, giveaways often
include terms like ``Wolfgang Mozart'' and ``Tamino'' that make
reaching an answer easier.
Consequently, our vector space \abr{ir} prove to be a strong baseline
(Section~\ref{sec:guess-eval}).

To frame this as an \abr{ir} search problem, we treat guessing as
document retrieval.
Input questions are search queries and embedded into a
\abr{tf-idf}~\citep{Jones1988ASI,rajaraman_ullman_2011} vector
$\bm{x}$.
For each answer $A_i\in\mathcal{A}_{train}$ in the \qb{} training
data, we concatenate all training questions with that answer into a
document $D_i$ embedded as $\bm{a}_i$ into the same vector
space.\footnote{ We also tested, one document per training example,
  different values for \abr{bm25} coefficients, and the default Lucene
  practical scoring function.}
The textual similarity function $f$ is
Okapi \abr{bm25}~\citep{RobertsonW94} and scores answers $\bm{a}_i$
against $\bm{x}$.
During inference, we return the answer $A_i$ of the highest scoring
document $D_i$.  We implement our model using Apache Lucene and
Elastic Search~\citep{gormley2015elasticsearch}.



However, the \abr{ir} model's reliance on pattern matching often fails
early in the question.
For example, in the first sentence from Figure~\ref{fig:ex} the author intentionally avoids keywords (``a character who asks for a glass of wine with his dying wish'').
Purely traditional \abr{ir} methods, while effective, are limited since they rely on keywords and cannot ``soft match'' terms semantically.
Thus, we move on to machine learning methods that address some of these shortcomings.

\subsection{Trainable Pattern Matching with Linear Models}
\label{sec:linear}

In addition to the \abr{ir} model, we also test a
linear model baseline that reduces multi-class classification
to one-versus-all binary classification.
While an \abr{ir} model derives term weights from corpus statistics
and a hand-crafted weighting scheme, a one-versus-all linear model
with one-hot term features $\bm{x}$ finds term weights that maximize
the probability of the correct binary prediction for each answer.
The input features $\bm{x}$ are derived from a combination of sparse
n-grams and skip-grams.\footnote{The order of n-grams and skip-grams
  was determined by hyper parameter search}
Since the number of classes is too high for standard one-versus-all
multi-class classification,\footnote{There are approximately 25,000
  distinct answers.} we instead use a logarithmic time one-versus-all
model~\citep{Agarwal2014ARE,daume16recalltree}.
However, this model is limited since it only considers linear
relationships between n-gram terms, the model uses--at best---local
word order, and the sparse representation does not take advantage of
the distributional hypothesis~\citep{harris1954distributional}.
Next we describe neural models that use more sophisticated forms of
representation and composition to address these shortcomings.

\subsection{Neural Network Models}
\label{sec:neural}

The final family of methods we consider for \qb{} question answering
are neural methods.
We describe the shared components of the neural models (e.g., general
architectures and training details) and compare their composition
functions.

\begin{figure*}
  \centering
  \includegraphics[width=.4\textwidth]{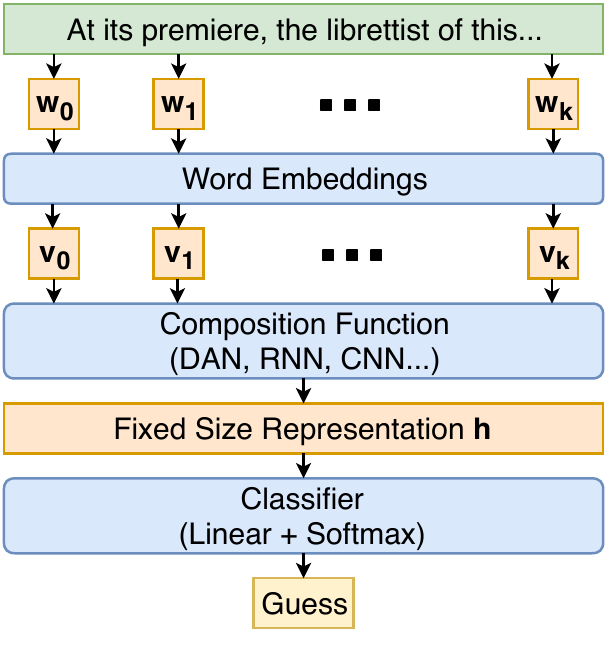}
  \caption{
    All our neural models feed their input to an embedding function,
    then a composition function, and finally a classification
    function.
    The primary variation across our models is the choice of
    composition function used to compute a fixed, example-level
    representation from its variable length input.  }
  \label{fig:nn}
\end{figure*}

In our model (Figure~\ref{fig:nn}), we follow a widely used
architecture in \abr{nlp} to embed words independently in a vector
space, contextualize their representations, temporally reduce
representations, and then classify with a softmax
layer~\citep{collobert2008unified}.
The first component of the model embeds question $q$ with $k$ tokens
into $m$-dimensional representations
$\bm{w}=[\bm{w}_1,\ldots,\bm{w}_k]$.  Next, a function $c(\cdot):
  \R^{k\times m}\rightarrow\R^{k\times l}$ contextualizes words as
$l$-dimensional embeddings
$\bm{v}=[\bm{v}_1,\ldots,\bm{v}_k]=c(\bm{w})$.
Since this is still a variable length sequence of representations and
the classifier requires a fixed size representation, we use a reducer
$r(\cdot): \R^{k\times m}\rightarrow\R^n$ to derive an $n$-dimensional
dense feature vector $\bm{x}=r(\bm{v})$.
We call specific pairs of contextualizers and reducers \textit{composition functions}.
The final model component---the classifier---computes logit scores $s_i=\bm{x}^T\cdot\bm{a}_i$ as the dot product between the features $\bm{x}$ and trainable answer embeddings $\bm{a}_i$.
From this, we use the softmax to compute a probability distribution
\begin{equation}
  \bm{p}=\text{softmax}(\bm{s})=\frac{\text{exp}(\bm{s})}{\sum_{i=1}^k \text{exp}(s_i)}
\end{equation}
over answers and train the model with the cross entropy loss
\begin{equation}
  \mathcal{L}=\sum_i^k y_i\log (\hat{p}_i)
\end{equation}
where $y_i=1$ for the true answer and $y_i=0$ otherwise.
In our experiments, we evaluate three classes of composition functions (i.e., contextualizer-reducer pairs): unordered composition with deep averaging networks~\citep{Iyyer:Manjunatha:Boyd-Graber:Daume-III-2015}, recurrent network-based composition~\citep{Elman1990FindingSI,DBLP:journals/neco/HochreiterS97,Palangi2016DeepSE,DBLP:conf/ssst/ChoMBB14}, and transformer-based composition~\citep{Vaswani2017AttentionIA,devlin2018bert}.

\subsubsection{Unordered Composition with Deep Averaging Networks}
\label{sec:dan}

Our first (unordered) neural composition function is the deep averaging network (\dan{}).
We introduced \dan{}s as a simple, effective, and efficient method for \qb{} question answering.\footnote{
This article has new experiments comparing new composition functions and focuses on incorporating additional data.
The \abr{dan} first was introduced in:\\ Mohit Iyyer, Varun Manjunatha, Jordan Boyd-Graber, and Hal
Daum\'{e} III.  {\bf
    \href{http://umiacs.umd.edu/~jbg//docs/2015_acl_dan.pdf}{Deep
      Unordered Composition Rivals Syntactic Methods for Text
      Classification}}.  \emph{Association for Computational
  Linguistics}, 2015.}
Despite their disregard of word order, \dan{}s are competitive with more sophisticated models on classification tasks such as sentiment analysis~\citep{Iyyer:Manjunatha:Boyd-Graber:Daume-III-2015}.
Although there are cases where word order and syntax matter, many questions are answerable using only key phrases.
For example, predicting the mostly likely answer to the bag of words ``inventor, relativity, special, general'' is easy; they are strongly associated with \underline{Albert Einstein}.

All composition functions---such as \dan{}s---are fully described by the choice of contextualizer and reducer.
In \dan{}s, the contextualizer $c$ is the identity function, and the reducer is broken into two components.
First, the \dan{} averages word embeddings $\bm{v}$ to create an initial hidden state
\begin{equation}
  \bm{h}_0=\frac{1}{k}\sum_{i=1}^k \bm{v}_i\text{.}
\end{equation}
The final fixed-size representation $\bm{x}=\bm{h}_z$ is computed with $z$ feed-forward layers through the recurrence
\begin{equation}
  \bm{h}_i=\textsc{gelu}(\bm{W}_i\cdot \bm{h}_{i-1}+\bm{b}_i)
\end{equation}
where $\bm{W}_i$ and $\bm{b}_i$ are parameters of the model and \textsc{gelu} is the Guassian Error Linear Unit~\citep{Hendrycks2016GaussianEL}.
Although \dan{}s are not the most accurate model, they are an attractive trade-off between accuracy and computation cost.

\subsubsection{Ordered Composition}
\label{sec:rnn}

In contrast to \dan{}s, order-aware models like \abr{rnn}s,
\abr{lstm}s, and \abr{gru}s can model long range dependencies in
supervised tasks~\citep{Linzen2016AssessingTA}.
Since all these models belong to the family of recurrent models, we
choose one variant to describe in terms of its associated
contextualizer and reducer.\footnote{ Hyper parameter optimization
  indicated that \gru{} networks were the most accurate recurrent
  model.  }
In our model, the composition function
\begin{equation}
  c(\bm{v})=\textsc{gru}(\bm{v})
\end{equation}
is a multi-layer, bi-directional \gru{}~\citep{DBLP:conf/ssst/ChoMBB14}.
The reducer
\begin{equation}
  r(\bm{v})=[\bm{v}_k^{(\text{forward})};\bm{v}_0^{(\text{backward})}]
\end{equation}
concatenates the final layer's forward and backward hidden states.
Combined, this forms the first ordered composition we test.

Transformer models, however, better represent context at the cost of
complexity~\citep{Vaswani2017AttentionIA,devlin2018bert}.
Specifically, we input the \abr{cls} token, question, and \abr{sep}
token to uncased \abr{bert-base}.  Thus, the contextualizer
\begin{equation}
  c(\bm{v})=\textsc{bert}(\bm{v})
\end{equation}
is simply \bert{} and the reducer
\begin{equation}
  r(\bm{v})=\frac{1}{k}\sum_{i=1}^k \bm{v}_i
\end{equation}
is the average of the output states from the final layer associated with the question's wordpiece tokens.\footnote{
  We also tested using the \abr{cls} token with worse results.
}
Next, we move on from model descriptions to training specifics.

\subsubsection{Training Details}
\label{sec:train-details}

In standard \qa{} tasks, training over full questions is standard,
but with \qb{}'s incremental setup this results in less accurate
predictions.
If the example is the complete text, the model ignores difficult clues
to focus on the ``easy'' part of the question, preventing learning
from ``hard'' clues.
Instead of each training example being one question, we use each of a
question's sentences as a singe training example.
While this training scheme carries the downside that during training
models may not learn long-range dependencies across sentences, the
accuracy improvement outweighs the disadvantages.
In addition to these two approaches, we also tested variable-length,
but did not observe an improvement over the sentence-based training
scheme.\footnote{ Variable length training creates $k$ training
  examples from a question comprised of $k$ sentences.  Each example
  includes the text from the start position up to and including
  sentence $k$.}

In non-transformer models we use 300-dimensional word embeddings
initialized with \glove{} for words in the vocabulary and randomly
initialized embeddings otherwise.\footnote{ Randomly initialized
  embeddings use a normal distribution with mean zero and standard
  deviation one.
}
We regularize these models with
dropout~\citep{DBLP:journals/jmlr/SrivastavaHKSS14} and batch
normalization~\citep{DBLP:conf/icml/IoffeS15}.
Loss functions were optimized with
\textsc{adam}~\citep{Kingma2014AdamAM} and models were trained with
early stopping, and learning rate annealing.
All neural models were implemented in PyTorch~\citep{NEURIPS2019_9015}
and AllenNLP~\citep{Gardner2017AllenNLP}.

We optimize hyper parameters by running each setting once and record
the parameter settings corresponding to the top development set accuracy.
The models with the best parameters are run an (additional) five times
to create estimates of variance for each tracked metric
(Section~\ref{sec:guess-eval}).

Although not exhaustive, these models are strong baselines for the
question answering component of \qb{}.
Section~\ref{sec:future-work} identifies areas for future modeling
work; throughout the rest of this work however we focus on completing
a description of our approach to playing \qb{} by combining guessers
and buzzer (Section~\ref{sec:buzzing}).
Following this we describe how we evaluate these systems independently
(Section~\ref{sec:guess-eval}), jointly (Section~\ref{sec:buzz-eval}),
offline (Section~\ref{sec:curve_score}), and live
(Section~\ref{sec:live}).

%% file: 2021_jmlr_qanta/sections/60-buzzing.tex
\section{Buzzing}
\label{sec:buzzing}

Winning \qb{} requires answering accurately with as little information
as possible.
It is crucial---for humans and computers alike---to accurately measure
their confidence and buzz as early as possible without being overly
aggressive.
The first part of our system, the guesser, optimizes for guessing
accuracy; the second part, the buzzer, focuses on deciding when to
buzz.
Since questions are revealed word-by-word the buzzer makes a binary
decision at each word: buzz and answer with the current best guess, or
wait for more clues.

The outcome of this action depends on the answers from both our
guesser and the opponent.\footnote{We use point values from the
  typical American format of the game.  The exact values are
  unimportant, as they change the particulars of strategy but not the
  approach.}
To make this clear, we review the game's mechanics.
If we buzz with the correct answer before the opponent can do so, we
win 10 points;
but if we buzz with an incorrect answer, we lose 5 points immediately,
and since we cannot buzz again, the opponent can wait till the end of
the question to answer, which might cost us 10 extra points in the
competition.

Before we discuss our strategy to buzzing, consider a buzzer
with perfect knowledge of whether the guesser is correct or not, but
does not know anything about the opponent: a \textit{locally optimal}
buzzer.
This buzzer would buzz as soon as the guesser gets the answer correct.
A stronger buzzer exists: an omnipotent buzzer with perfect knowledge
of what the opponent will do; it would exploit the opponent's
weaknesses: delay buzzing whenever an opponent might err.
The agent would then get a higher relative reward: once from the
opponent's mistake and then for getting it correct.

The buzzer we develop in this paper targets a locally optimal
strategy: we focus on predicting the correctness of the guesser and do
not model the opponent.
This buzzer is effective: it both defeats players in our gameplay
dataset (Section~\ref{sec:protobowl-data}) and playing against real
human players (Section~\ref{sec:live}).
The opponent modeling extension has been explored by previous work,
and we discuss it in Section~\ref{sec:rel}.

\subsection{A Classification Approach to Buzzing}

%
%
%

Given the initial formulation of buzzing as a \abr{mdp} (Section~\ref{sec:form-task}), it would be natural to learn the task with reinforcement learning using the final score; however, we instead use a convenient reduction to binary classification.
Since we can compute the optimal buzzing position easily as opposed to with expensive rollouts, we can reduce the problem to classification~\citep{lagoudakis2003class}.
At each time step, the model looks at the sequence of guesses that the
guesser has generated so far, and makes a binary decision of whether to
buzz or to wait.
Under the locally optimal assumption, the ground truth action at each
time step equals the correctness of the top guess: it should buzz if
and only if the current top guess is correct.
Another view of this process is that the buzzer is learning to imitate the oracle buzzing policy from the ground truth actions~\citep{coates2008imitation,ross2010reduction,ross2011noregret}.
Alternatively, the buzzer can also be seen as an uncertainty
estimator~\citep{hendrycks2016baseline} of the guesser.

The guesses create a distribution over all possible answers.
If this distribution faithfully reflects the uncertainty of guesses,
the buzzer could be a simple ``if--then'' rule: buzz as soon as the
guesser probability for any guess gets over a certain threshold.
This \emph{threshold} system is our first baseline, and we tune the
threshold value on a held-out dataset.

However, this does not work because the confidence of neural models is
ill-calibrated~\citep{Guo:Pleiss:Sun:Weinberger-2017,feng2018rawr}.
Our neural network guesser often outputs a long tail distribution over
answers concentrated on the top few guesses, and the confidence score
of the top guess is often higher than the actual uncertainty (the
chance of being correct).
To counter these issues, we extract features from the top ten guesser
scores train a classifier on top of them.
Some important features include a normalized version of the top ten
scores and the gap between them; the full list of features are in Appendix~\ref{apx:buzzer_features}.

There is also important temporal information; for example, if the
guesser's top prediction's score steadily increases, this signals the
guesser is certain about the top guess.
Conversely, a fluctuating top prediction (the answer is \underline{Hope
  Diamond}\dots no, I mean \underline{Parasaurolophus}\dots no, I mean
\underline{Tennis Court Oath}) is a sign that perhaps the guesser is
not that confident (regardless of the ostensible score).
To capture this, we compare the current guesser
scores with the previous time steps and extract features such as the
change in the score associated with the current best guess, and
whether the ranking of the current top guess changed in this time
step.
The full list of temporal features can be found in the
Appendix~\ref{apx:buzzer_features}.

To summarize, at each time step, we extract a feature vector,
including current and temporal features, from the sequence of guesses
generated by the guesser so far.
We implement the classifier with both fully connected Multi-layer
Perceptron (\abr{mlp}) and with Recurrent Neural Network (\abr{rnn}).
The classifier outputs a score between zero and one indicating the
estimated probability of buzzing.
Following the locally optimal assumption, we use the correctness of
the top guess as ground truth action: buzz if correct and wait if
otherwise We train the classifier with logistic regression; during
testing, we buzz as soon as the buzzer outputs a score greater than
$0.5$.
Both models are implemented in
Chainer~\citep{chainer_learningsys2015}; we use hidden size of $100$,
and \abr{lstm} as the recurrent architecture.
We train the buzzer on the ``buzzertrain'' fold of the dataset, which
does not overlap with the training set of the guesser, for twenty epochs
with Adam optimizer~\citep{Kingma2014AdamAM}.
Both buzzers have test accuracy of above 80\%, however, the
classification accuracy does not directly translate into the buzzer's
performance as part of the pipeline, which we look at next.

%% file: 2021_jmlr_qanta/sections/65-eval.tex
\section{Offline Evaluation}
\label{sec:eval}

A central thesis of our work is that the construction of \qb{}
questions lends itself to a fairer evaluation of both humans and
machine \abr{qa} models: to see who is better at answering questions,
see who can answer the question first.
However, this is often impractical during model development,
especially if the questions are ``new'' (they have not been played by
humans or computers).
Moreover, a researcher might be uninterested in solving the buzzing
problem.
Offline evaluations where the guesser and buzzer are evaluated independently with static data strikes a balance between ease of model development and faithfulness to \qb{}'s format.
To address this, Section~\ref{sec:guess-eval} describes the metrics to compare offline model accuracy.
Following an error analysis (Section~\ref{sec:error}),
Section~\ref{sec:buzz-eval} evaluates buzzing models by replacing this
oracle buzzer with trained buzzing models.

\subsection{Evaluating the Guesser}
\label{sec:guess-eval}

Ideally, we would compare systems in a head-to-head competition where
the model (or human) who correctly buzzed and answered the most
questions would win (Section~\ref{sec:live}).
However, this involves live play, necessitates a buzzing strategy, and
complicates evaluation of the guesser in isolation.
Intuitively though, a model that consistently buzzes correctly earlier
in the question is better than a model that buzzes late in the
question.
In our evaluations, we use three metrics that reflect this intuition:
accuracy early in the question, accuracy late in the question, and the
expected probability of beating a human assuming an optimal buzzing
strategy.

\subsubsection{Accuracy-Based Evaluation}

The easiest and most common method for evaluating closed domain
question answering methods is accuracy over all questions in the test
set.
We report two variants of this: (1) accuracy using the first sentence
and (2) accuracy using the full question.
While it is possible to answer some questions during the first
sentence, it is the first and hardest position we can guarantee
\emph{could} be answered.
Although we report accuracy on full questions, this metric is a
minimum bar: the last clues are intentionally easy
(Section~\ref{sec:craft}).
However, while start-of-question and end-of-question accuracy help
development and comparison with other \qa{} tasks, it is silent on
human--computer comparison.
We address this shortcoming next.

\subsubsection{Expected Probability of Defeating Human Players}
\label{sec:curve_score}

While comparing when systems buzz is the gold standard,
we lack gameplay records for all test set questions, and it is
unreasonable to assume it is easy to obtain them.
Instead, marginalize over empirical human gameplay to estimate the
probability $\pi(t)$ that a human would have correctly answered a
question by position $t$.
Then, we combine this with model predictions and marginalize over $t$
to obtain the expected probability of winning against an average
player on an average gameplay question.
A similar idea---to compute the expected probability of winning a heads up match---has also been used in machine translation~\citep{bojar2013wins}.

\begin{figure}[t]
  \centering
  \includegraphics[width=\linewidth]{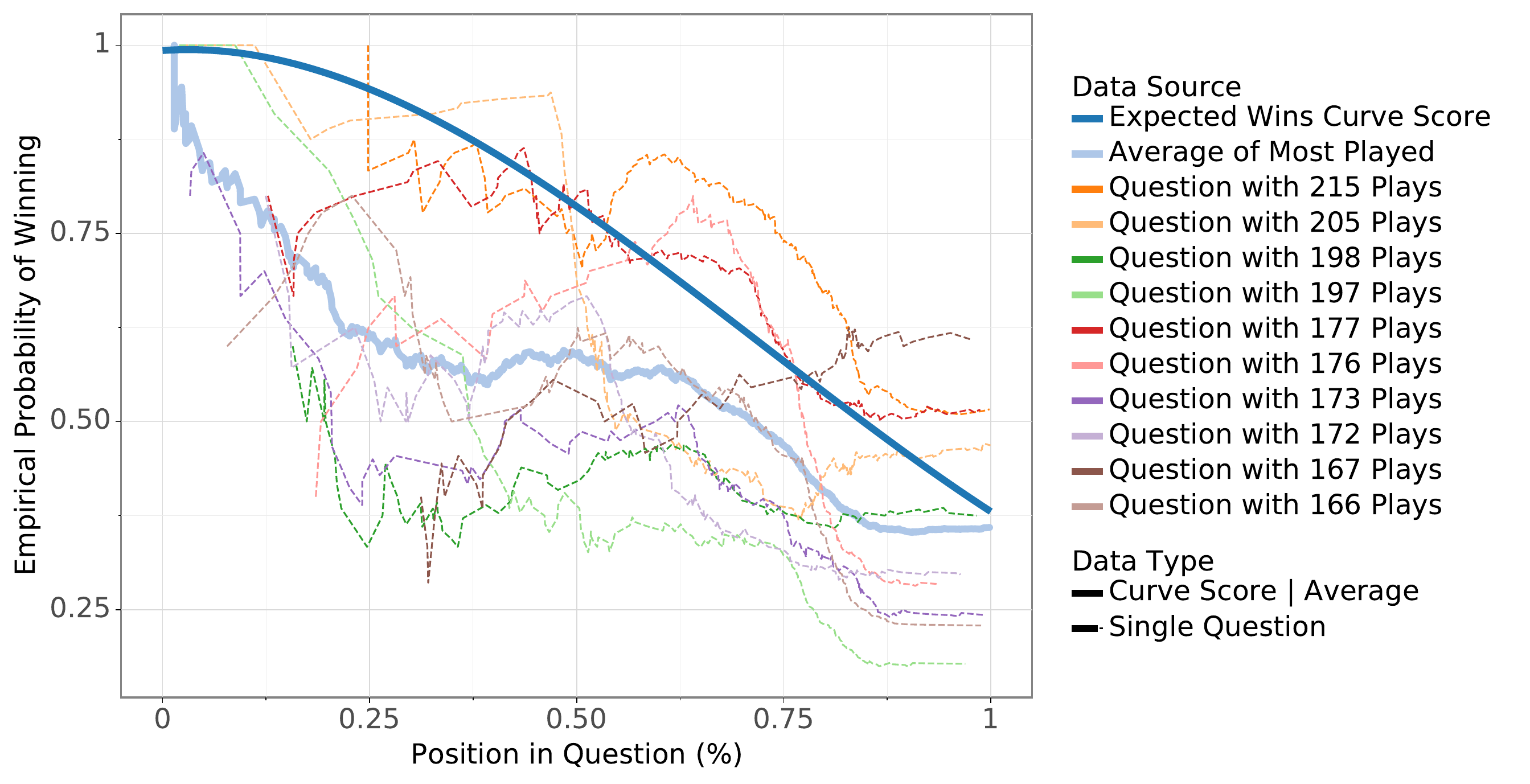}
  \caption{
    We plot the expected wins score with respect to buzzing position (solid dark blue).
    For the ten most played questions in the buzztest fold we show the empirical distribution for each individual question (dotted lines) and when aggregated together (solid light blue).
    Among the most played questions, expected wins over-rewards early buzzes, but appropriately rewards end-of-question buzzes.
  }
  \label{fig:empirical-ew}
\end{figure}

We compute the \textbf{e}xpected probability of \textbf{w}inning
(\abr{ew}) in two steps.
First, we compute the proportion of players
\begin{equation}
  \label{eq:emp-w}
  \pi(t)=1-\frac{N_t}{N},
\end{equation}
that have answered a question correctly by position $t$.
$N$ is the total number of question-player records and $N_t$ is the
number of question-player records where the player answered correctly
by position $t$.
We empirically estimate the expected probability of winning
\begin{equation}
  \label{eq:ew-cubic}
  \pi(t)= 0.0775t - 1.278t^2 + 0.588t^3
\end{equation}
from the gameplay data as a cubic polynomial (Figure~\ref{fig:empirical-ew}).
At $t=0$, the potential payoff is at its highest since no one has
answered the question.
At $t=\infty$, the potential payoff is at its lowest; all the players
who would have correctly answered the question already have.
If the computer gets the question right at the end, it would only score points against opponents who did not know the answer at all or answered incorrectly earlier in the question.

\abr{ew} marginalizes over all questions~$q$ and all positions~$j$,
and counts how many times model~$m$ produced a guess~$g(m,q,j)$ that
matched the answer of the question~$a(q)$.
Specifically we compute
\begin{align}
  \label{eq:ew}
  \mbox{EW}(m) & =\e{m}{p_{\text{win}}} =\frac{1}{|\mathcal{Q}|}\sum_{q\in \mathcal{Q}}\sum_{j=1}^\infty
  \ind{g(m,q,j)=a(q)} \pi\left(j\right),
\end{align}
where $\frac{1}{|\mathcal{Q}|}$ is the count of question--position records.
The indicator function is exactly an oracle buzzer: it gives credit if and only if the answer is correct.
However, this rewards models with unstable predictions; for example, a
model would be rewarded twice for a sequence of predictions that were
correct, wrong, and correct.
We discourage this model behavior by using a \textit{stable} variant
of \abr{ew} which only awards points if the current answer and all
subsequent answers are correct.
With this formalism, it is also straightforward to compute the
expected winning probability for any guesser-buzzer combination by
replacing the oracle buzzer (indicator function) with a function that
equals one if the guess is correct and the buzzer yielded a ``buzz''
decision.
We compare buzzers in Section~\ref{sec:buzzing}, but now move to
experimental results for the guesser.

\subsubsection{Guesser Comparison Experiments}

We evaluate our guessers using start accuracy, end accuracy, and
expected wins (Table~\ref{table:scores}).
All models struggle at the start of the question with the best
accuracy at only 11\%.
This is unsurprising: the first sentence contains the most difficult
clues and is difficult for even the best human players.
Models fare significantly better near the end of the question with
giveaway clues.
However, even the best model's 61\% accuracy leaves much room for
future work.

\begin{table}[t]
  \centering
  \small
  \include{2021_jmlr_qanta/auto_fig/experiment_table}
  \caption{
    We compare several models by accuracy at start-of-question, end-of-question, and \abr{ew}.
    In the table, models are sorted by start-of-question development set accuracy.
    Standard deviations for non-\abr{ir} models are derived from five trials; standard deviation is not reported for the \abr{ir} model since it is deterministic.
  }
  \label{table:scores}
\end{table}

While the \bert{} model has the best early-question accuracy, it lags
behind the \abr{ir} and \abr{dan} for end of question accuracy.
We suspect that order-aware models over-emphasize less important parts
of the question; additionally, the gap between sentence training and
full question inference advantages models that did not need
to learn an aggregation over longer sequences.
This pattern is also reflected in the \abr{ew} scores;\bert{}---as
expected---outperforms the \abr{rnn} model.
Finally, across accuracy and \abr{ew} we see substantial drops between
the development and test sets, which suggests overfitting.
Next, we investigate the errors models make.

\subsection{Identifying Sources of Error}
\label{sec:error}

This section identifies and characterizes several failure modes of our
models.
First we compare the predictions of blackbox neural models and
\abr{ir} model---an explicit pattern matcher
(Section~\ref{sec:cross-error}).
Following this we identify data skew towards popular answers as a
major source of error for less popular answers
(Section~\ref{sec:divergence}).
Lastly, we manually break down the test errors of one model
(Section~\ref{sec:breakdown}).

\subsubsection{Behavioral Comparison of Neural and IR Models}
\label{sec:cross-error}


\begin{figure}[t]
  \centering
  \begin{subfigure}{.45\textwidth}
    \centering
    \includegraphics[width=\linewidth]{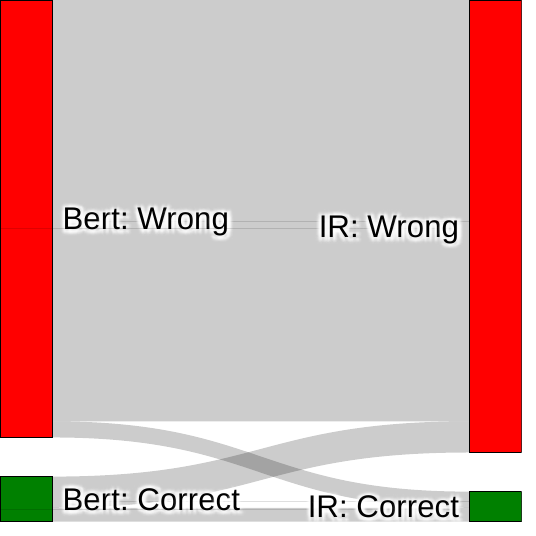}
    \caption{Start of the question error comparison}
    \label{fig:error-start}
  \end{subfigure}
  \begin{subfigure}{.45\textwidth}
    \centering
    \includegraphics[width=\linewidth]{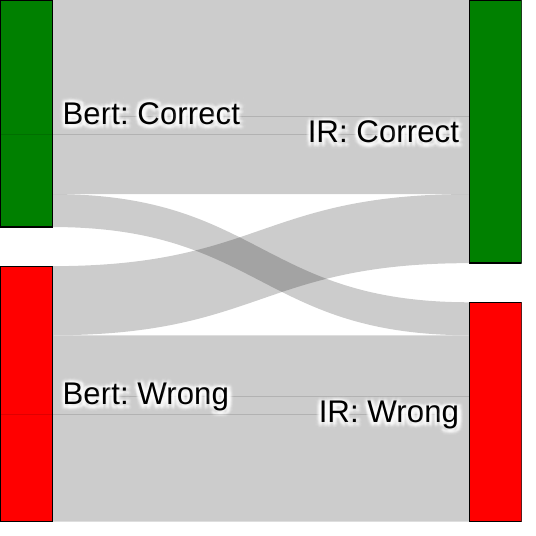}
    \caption{End of the question error comparison}
    \label{fig:error-end}
  \end{subfigure}
  \caption{
    The \bert{} and \abr{ir} models are mostly wrong or correct on the same subset of questions.
    At the end of the question, most of the questions the \bert{} model is correct on, the \abr{ir} model is also correct on.
  }
  \label{fig:guesser_comp}
\end{figure}

One way to analyze black-box models like neural networks is to compare
their predictions to better understood models like the \abr{ir} model.
If their predictions---and thus exterior behavior---are similar to a
better understood model it suggests that they may operate similarly.
Figure~\ref{fig:guesser_comp} shows that even the \bert{} and \abr{ir}
models are correct and wrong on many of the same examples at
end-of-question.
Since one model---the \abr{ir} model---is an explicit pattern matcher
this hints that neural \qb{} models learn to be pattern matcher as
hinted by other
work~\citep{Jia2017AdversarialEF,Rajpurkar2018KnowWY,feng2018rawr}.

\begin{figure}[t]
  \center
  \tikz\node[draw=black!40!lightblue,inner sep=1pt,line width=0.3mm,rounded corners=0.1cm]{
    \begin{tabular}{p{.9\textwidth}}
      \textbf{Test Question (first sentence)}:                                                                                           \\
      \hl{A holder of this title commissioned} a set of \hl{miniatures} to accompany the story collection Tales of a Parrot.             \\
      \textbf{Training Question (matched fragment)}:                                                                                     \\
      \hl{A holder of this title commissioned} Abd al-Samad to work on \hl{miniatures} for books such as the Tutinama and the Hamzanama. \\
      \textbf{Answer}: \underline{Mughal Emperors}
    \end{tabular}
  };
  \caption{A \href{http://datasets.pedro.ai/qanta/question/106373}{test question} that was answered correctly by all
    models after the first sentence; a normally very difficult task for both
    humans and machines. A \href{http://datasets.pedro.ai/qanta/question/118623}{very similar training example} allows all models
    to answer the question through trivial pattern matching.}
  \label{fig:success}
\end{figure}

Next we investigate this pattern matching hypothesis at the
instance-level.
For our instance-level analysis we sample examples of correct and
incorrect predictions.
First we randomly sample a test question that all models answer
correctly after the first sentence (Figure~\ref{fig:success}).
This particular example has similar phrasing to a training example
(``A holder of this title commissioned\dots miniatures'') so it is
unsurprising that all models get it right.

\begin{figure}[t]
  \begin{center}
    \tikz\node[draw=black!40!lightblue,inner sep=1pt,line width=0.3mm,rounded corners=0.1cm]{
      \begin{tabular}{p{.9\textwidth}}
        \textbf{Test Question (first sentence)}:                                  \\
        This phenomenon is resolved without the help of a theoretical model in
        costly DNS methods, which numerically solve for the rank-2 tensor
        appearing in the RANS equations.                                          \\
        \textbf{Answer}: \underline{Turbulence} \textbf{Score} (\abr{rnn}): .0113 \\
        \textbf{Synonym Attacks}: phenomenon $\rightarrow$ event, model $\rightarrow$ representation
      \end{tabular}
    };
  \end{center}
  \caption{
    Only the \abr{rnn} model answers \href{http://datasets.pedro.ai/qanta/question/106253}{this question} correctly.
    To test the robustness of the model to semantically equivalent input modifications, we use \abr{sears}-based~\citep{Singh2018SemanticallyEA} synonym attacks and cause the model prediction to become incorrect.
    Although this exposes a flaw of the model, it is also likely that the low confidence score would likely lead a buzzer model to obstain; this highlights one benefit of implicitly incorporating confidence estimation into the evaluation.
  }
  \label{fig:success-rnn}
\end{figure}

In our second analysis, we focus on a specific answer
(\underline{Turbulence}) and its twenty-seven training questions.
Figure~\ref{fig:success-rnn} shows a sample question for this answer
that the \abr{rnn} model answered correctly but that \abr{ir} model
did not.
The most frequent words in the training data for this answer are
``phenomenon'' (twenty-three times), ``model'' (seventeen times),
``equation'' (thirteen times), ``numerically'' (once), and ``tensor''
(once).
In this analysis we removed or substituted these word with synonyms
and then checked if the model's prediction was the same.

Substituting words in this question shows that the model is
over-reliant on specific terms.
After removing the term ``phenomenon,'' the model changed its answer
to \underline{Ising model} (a mathematical model of ferromagnetism).
If we instead substitute the term with synonyms such as
``occurrence'', ``event'', and ``observable event'' the answers are
still incorrect.
Similarly, if ``model'' is replaced by ``representation'' the
\abr{rnn} also makes incorrect predictions.
At least for this question, the model is not robust to these
semantics-preserving modifications~\citep{Singh2018SemanticallyEA}.
Next we move to aggregate error analysis.

\subsubsection{Errors Caused by Data Sparsity}
\label{sec:divergence}

\begin{figure*}[t]
  \centering
  \includegraphics[width=\textwidth]{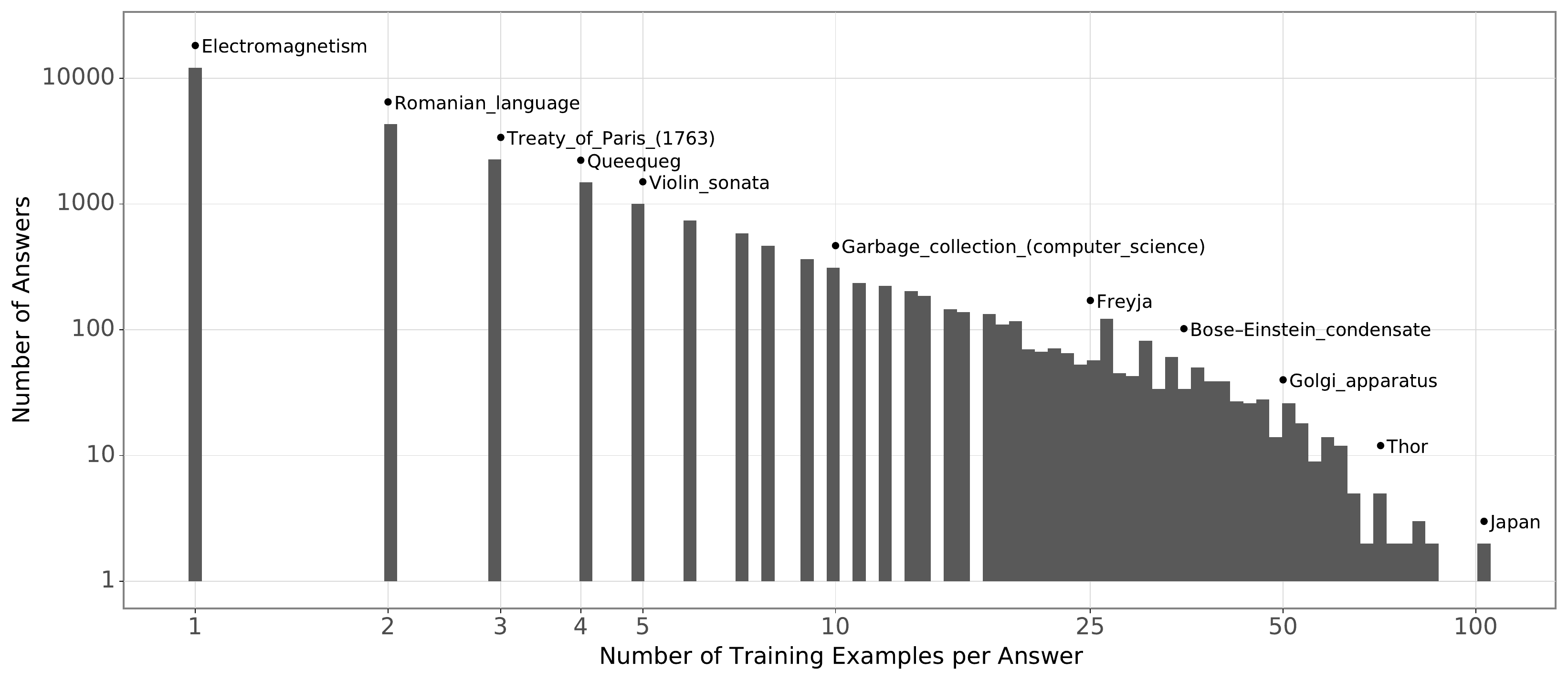}
  \caption{
    The distribution of training examples per unique answer is heavily skewed.
    The most frequent answer (Japan) occurs about 100 times.
    Nearly half of the questions have one training example and just over sixty percent have either one or two training examples.
  }
  \label{fig:ntrain-cdf}
\end{figure*}

\begin{figure*}[t]
  \centering
  \includegraphics[width=\textwidth]{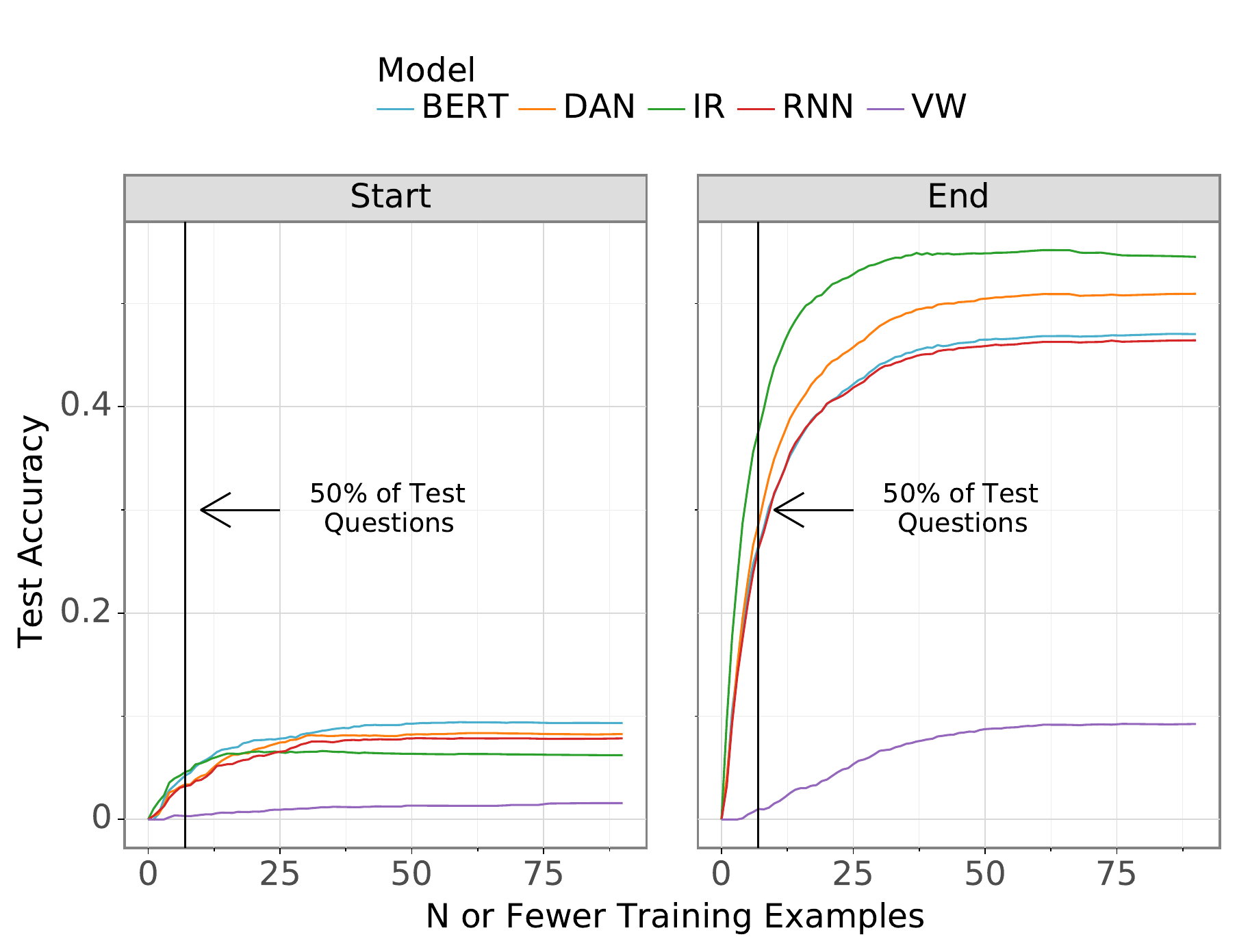}
  \caption{
    The more an answer is asked in the training set, the easier it is
    for all models, both at the start and end of the question.
    This is a significant source of errors since accuracy on at least 50\% of test
    questions---those with seven or less training examples---is significantly lower
    for all models.
  }
  \label{fig:acc_ntrain}
\end{figure*}

For many test set answers, scarcity of training data is a significant
source of error.
Most egregiously, $17.9\%$ of test questions have zero corresponding
training examples.
Beyond these questions, many more answers have few training examples.
While some topics are frequently asked about, one goal of question
writers is to introduce new topics for students to learn from.
For example, although physics is a common general topic,
\underline{Electromagnetism} has only been an answer to one \qb{}
question.
The distribution of training examples per unique answers is skewed
(Figure~\ref{fig:ntrain-cdf}), and countries---like
\underline{Japan}---are asked about much more frequently.
Unsurprisingly, plotting the number of training examples per test
question answer versus model accuracy shows significant drops in
accuracy for about half of the test questions
(Figure~\ref{fig:acc_ntrain}).

\subsubsection{Error Breakdown}
\label{sec:breakdown}

\begin{table}[t]
  \centering
  \begin{tabular}{l r r l}
    \toprule
    Error Reason  & Start Count & End Count \\
    \midrule
    Wrong Country & 11          & 17        \\
    Wrong Person  & 16          & 2         \\
    Wrong Place   & 1           & 5         \\
    Wrong Type    & 15          & 5         \\
    Wrong Event   & 0           & 1         \\
    Nonsense      & 7           & 2         \\
    Annotation    & 1           & 4         \\
    \bottomrule
  \end{tabular}
  \caption{
    The table is an error breakdown for questions with at least twenty-five training examples.
    To analyze errors at the start of questions, we randomly sampled fifty errors and for end of question took all thirty-six errors.
    End of question errors are primarily wrong country errors as in Figure~\ref{fig:wrong-country} where the model answers \underline{United States} instead of \underline{Spain}.
    Errors at the start of the question though are more diverse.
    The most common error is guessing the correct answer type, but not the specific member of that type; examples of this error class include answering \underline{Albert Einstein} instead of \underline{Alan Turing}, or \underline{Iowa} instead of \underline{Idaho}.
  }
  \label{table:errors}
\end{table}

We conclude our error analysis by inspecting and breaking down the
errors made by the \abr{rnn} model at the start and end of questions.
Of the $2,151$ questions in the test set, $386$ have zero training
examples leaving $1,765$ questions that are answerable by our models.
Of the remaining questions, the \abr{rnn} answers $1,540$ incorrectly
after the first sentence and $481$ at the end of the question.
To avoid errors likely due to data scarcity, we only look to questions
with at least $25$ training examples; the number of errors on this
subset at the start and the end of the question is $289$ and $36$.
Table~\ref{table:errors} lists reasons for model errors on a random
sample of $50$ errors from the start of the question and all $36$
errors from the end of the question.

The predominant sources of error are when the model predicts the
correct answer type (e.g., person, country, place), but chooses the
incorrect member of that type.
This accounts for errors such as choosing the wrong person, country,
place, or event.
The \abr{rnn} especially confuses countries; for example, in
Figure~\ref{fig:wrong-country}, it confuses the \underline{Spain} and
the \underline{United States}, the parties to the Adam Onis Treaty.
The relative absence of incorrect answer type errors at the end of
questions may be attributable to the tendency of late clues including
the answer type (such as ``name this country\ldots'').

\begin{figure}[ht]
  \center
  \tikz\node[draw=black!40!lightblue,inner sep=1pt,line width=0.3mm,rounded corners=0.1cm]{
    \begin{tabular}{p{.9\textwidth}}
      \textbf{Test Question}: This country seized four vessels owned by Captain
      John Meares, which were registered in Macau and disguised with Portuguese
      flags, starting a dispute over fishing rights. To further negotiations
      with this country, Thomas Jefferson signed the so-called ``Two Million
      Dollar Act.'' This country agreed not to police a disputed spot of land,
      which was subsequently settled by outlaws and ``Redbones'', and which was
      called the ``Neutral Ground.'' This country was humiliated by England in
      the Nootka Crisis. Harman Blennerhassett's farm on an island in the Ohio
      River was intended as the launching point of an expedition against this
      European country's possessions in a plan exposed by James Wilkinson. This
      country settled navigation rights with the United States in Pinckney's
      Treaty, which dealt with the disputed ``West'' section of a colony it
      owned. For 10 points, name this European country which agreed to the
      Adams-Onis Treaty handing over Florida.                                      \\
      \textbf{Guess}: \underline{United States} \textbf{Answer}: \underline{Spain} \\
    \end{tabular}
  };

  \caption{
    Although the answer to this question is \underline{Spain}, many of the terms and phrases mentioned are correlated with the \underline{United States}.
    Thus, the \abr{rnn} model answers \underline{United States} instead of the correct answer \underline{Spain}.
    This is one of many examples where the model answers with the correct answer type (country), but incorrect member of that type.
  }
  \label{fig:wrong-country}
\end{figure}

In the process of manual error breakdown we also found five annotation
errors where the assigned Wikipedia answer did not match the true
answer.
This low number of errors further validates the robustness of our
answer mapping process.

\subsection{Evaluating the Buzzer}
\label{sec:buzz-eval}

We first evaluate our buzzer against the locally optimal buzzer which
buzzes as soon as the guesser gets the answer correct.
However, this can be overly ambitious and unrealistic since the
guesser is not perfectly stable: it can get the answer correct by
chance, then vacillate between several candidates before settling down
to the correct answer.
To account for this instability, we find the first position that the
guesser stabilizes to the correct answer and set it as the optimal
buzzing position. In other words, we compare against the optimal
buzzer used in the definition of stable \abr{ew} score.
To be exact, we start at the last position that the
guess is correct, go backwards until the guess is incorrect and
consider this the locally optimal buzzing position; we set the
ground truth to all positions before this to zero, and all positions
after it to one.

We use the same guesser (\abr{rnn}) in combination with different
buzzers (Table~\ref{table:buzzers_ew}), and quantitatively compare
their expected wins (Section~\ref{sec:curve_score}).
Both \abr{mlp} and \abr{rnn} buzzers win against the static threshold
baseline, but there is a considerable gap
between \abr{rnn} and the optimal buzzer.

\begin{figure}[ht]
  \begin{center}
    \tikz\node[draw=black!40!lightblue,inner sep=1pt,line width=0.3mm,rounded corners=0.1cm]{
      \begin{tabular}{p{.9\textwidth}}
        This instrument plays the \ttriangle{cred} only extended solo in the overture to Verdi's \tdiamond{cred}$^1$ Luisa Miller. This is the solo instrument in a piece \tsquare{cred} that opens with the movement ``The Perilous Shore''. \diabox1{cblue} This instrument introduces \tcircle{cgreen} the main theme to ``The Pines of Janiculum'' from Respighi's The Pines of Rome. This instrument has a long solo at the beginning of the Adagio from Rachmaninoff's Second Symphony, and it first states the Shaker theme in Copland's Appalachian Spring. John Adams' Gnarly Buttons is for this instrument. Heinrich \tdiamond{cgreen} Baermann, a virtuoso on this instrument, was the \tdiamond{cgreen} dedicatee of Carl Maria von Weber's two concertos for it. The basset \tdiamond{cgreen} horn is a variant of, for 10 points, what \tdiamond{cred}$^2$ \tdiamond{cgreen} single-reed woodwind instrument, which plays \tdiamond{cgreen} a notable glissando at the opening of Gershwin's \tdiamond{cred}$^3$ Rhapsody in Blue? \\
        \textbf{Answer:} \underline{Clarinet} \textbf{Optimal Buzz:} \diabox1{cblue} \textbf{Correct:} \tdiamond{cgreen} \ttriangle{cgreen} \tcircle{cgreen} \tsquare{cgreen} \textbf{Wrong:} \tdiamond{cred} \ttriangle{cred} \tcircle{cred} \tsquare{cred}                                                                                                                                                                                                                                                                                                                                                                                                                                                                                                                                                                                                                                                                                                                                                                                      \\
        \textbf{Threshold Buzz:} \ttriangle{cred} \underline{Bassoon} \textbf{MLP Buzz:} \tsquare{cred} \underline{Bassoon} \textbf{RNN Buzz:} \tcircle{cgreen}  \underline{Clarinet}                                                                                                                                                                                                                                                                                                                                                                                                                                                                                                                                                                                                                                                                                                                                                                                                                                                             \\
        \textbf{Human Buzzes:} \tdiamond{cred}$^1$ \underline{Violin}, \tdiamond{cred}$^2$ \underline{Obo}, \tdiamond{cred}$^3$ \underline{Flute}, \tdiamond{cgreen} \underline{Clarinet}
      \end{tabular}
    };
  \end{center}
  \caption{
    In this question, the Threshold \ttriangle{cred} and \abr{mlp} \tsquare{cred} buzzers are too aggressive and buzz before the guesser's answer is correct.
    In contrast, the \abr{rnn} \tcircle{cgreen} is more conservative and buzzes shortly after the optimal point \diabox1{cblue} which is---by a wide margin---still earlier than the earliest (correct) human buzz \tdiamond{cgreen}.
  }
  \label{fig:ex-w-buzzes}
\end{figure}

\begin{figure}[t]
  \centering
  \includegraphics[width=.95\textwidth]{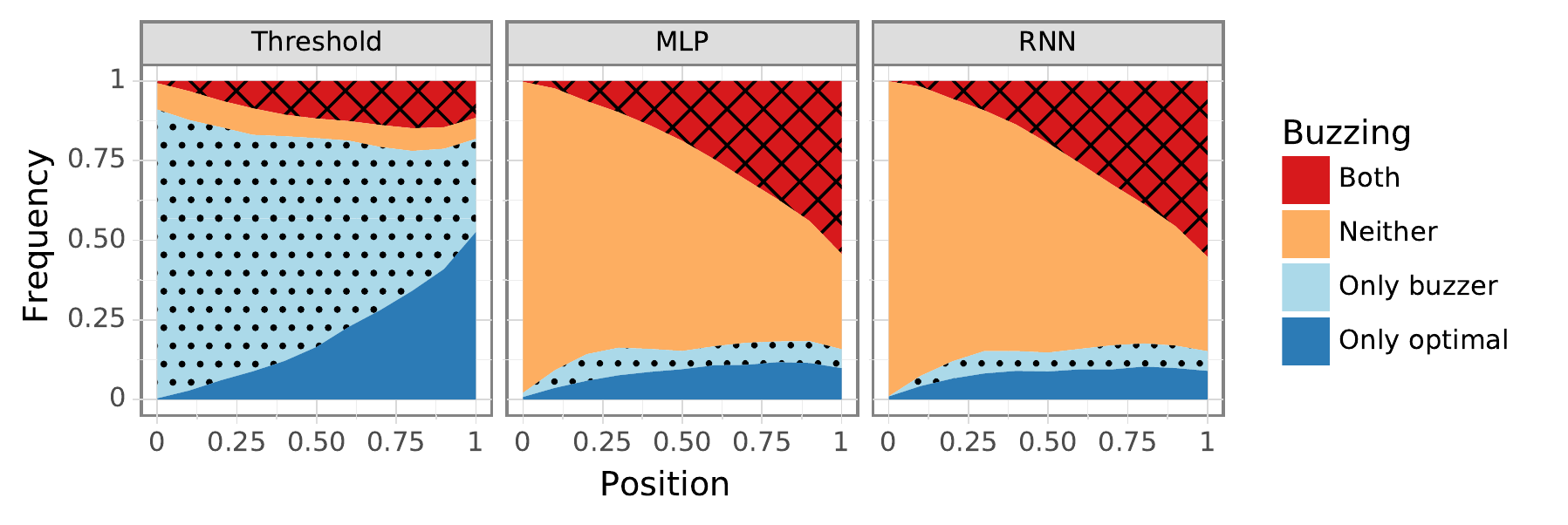}
  \caption{Comparing buzzers' behavior over time against the optimal
    buzzer. The red crossed area and dotted blue area combined indicates
    when the buzzer thinks that the guesser is correct, the other
    two combined when the buzzer thinks the guesser is wrong.
    The red (crossed) and orange (unhatched) areas combined indicates when
    the buzzer matches the optimal buzzer.
    Our goal is to maximize the red areas and minimize the blue areas.
    The static threshold baseline is overly aggressive, especially at
    earlier positions in the question (large dotted blue area);
    \abr{mlp} and \abr{rnn} both behaves reasonably well, and the
    aggressiveness of \abr{rnn} is slightly more balanced early on in the question.
  }

  \label{fig:buzzdev_stack_area}
\end{figure}

Low expected wins means the
buzzer is either too aggressive or not aggressive enough.
To characterize their weaknesses, we compare the buzzers' behavior
over time (Figure~\ref{fig:buzzdev_stack_area}).
The static threshold buzzer is too aggressive, especially early in the
questions as is also seen in Figure~\ref{fig:ex-w-buzzes}.
This behavior to some extent resonates with the observation that the
confidence of neural models needs
calibration~\citep{Guo:Pleiss:Sun:Weinberger-2017}.
The difference between \abr{mlp} and \abr{rnn} is small but \abr{rnn}
is less likely to be overly aggressive early in the question.

\begin{table}[t]
  \centering
  \begin{tabular}{cccc}
    \toprule
    Model     & \abr{acc} & \abr{ew} & Score \\\midrule
    Threshold & \         & 0.013    & -9.98 \\
    \abr{mlp} & 0.840     & 0.272    & -2.31 \\
    \abr{rnn} & 0.849     & 0.302    & -1.01 \\
    Optimal   & 1.0       & 0.502    & 2.19  \\
    \bottomrule
  \end{tabular}
  \caption{The accuracy (\abr{acc}), expected wins (\abr{ew}), and \qb{} score
    (Score) of each buzzer on the validation set. Both \abr{mlp} and
    \abr{rnn} outperform the static threshold baseline by a large margin,
    but there is still a considerable gap from the optimal buzzer.
  }
  \label{table:buzzers_ew}
\end{table}

\begin{figure}[t]
  \centering
  \includegraphics[width=.95\textwidth]{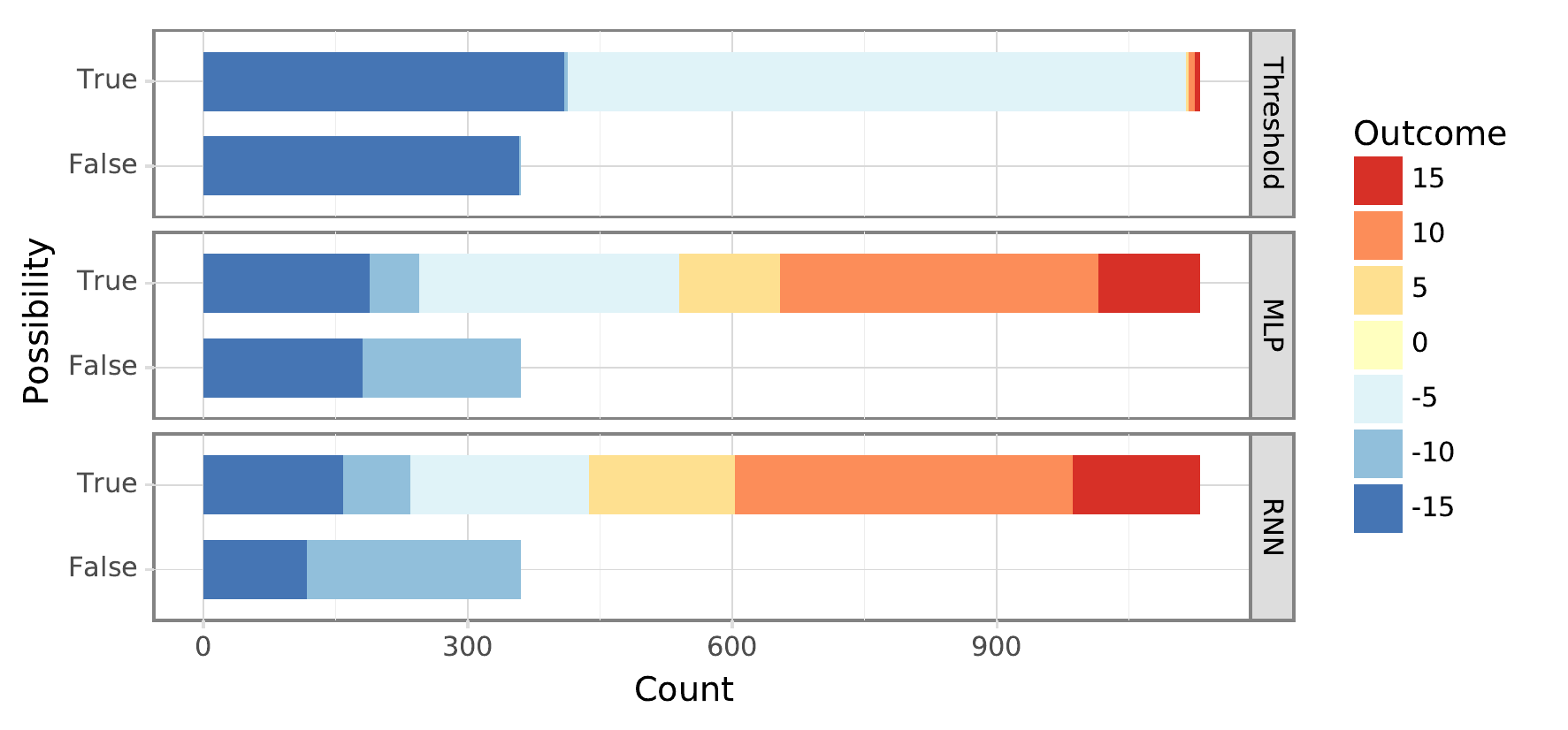}
  \caption{Breaking down the buzzer's performance on the individual
    question level. Impossible question means there is nothing the buzzer
    can do to beat the opponent. It is clearer that \abr{rnn} performs
    better than \abr{mlp}, making fewer mistakes of being overly
    aggressive.}
  \label{fig:buzzdev_protobowl}
\end{figure}

For a more fine-grained analysis, we simulate games where our system
plays against individual human players using the gameplay dataset
(Section~\ref{sec:protobowl-data}).
Based on the guesser, we classify questions as ``possible'' or not.
If the guesser gets the
answer correct before the opponent answers, it is \emph{possible} for
the buzzer to win the question.
Otherwise, it is impossible for the buzzer to do anything to
beat the opponent.
Based on this categorization, Figure~\ref{fig:buzzdev_protobowl}
further breaks down the outcomes: the
\abr{rnn} is less likely to be overly aggressive in both possible and
impossible cases.

%% file: 2021_jmlr_qanta/auto_fig/experiment_table.tex
	\begin{tabular}{ l r r r r r r r r r r}
		\toprule
    & \multicolumn{8}{c}{Accuracy (\%)} & &\\
    \cmidrule{3-10}
    & \multicolumn{4}{c}{Start} & \multicolumn{4}{c}{End} & &\\
    \cmidrule(lr){2-5}\cmidrule(lr){6-9}
    &
    \multicolumn{2}{c}{Dev} &
    \multicolumn{2}{c}{Test} &
    \multicolumn{2}{c}{Dev} &
    \multicolumn{2}{c}{Test} &
    \multicolumn{2}{c}{$\e{}{p_{\text{win}}}$}\\
    \cmidrule(lr){2-3}\cmidrule(lr){4-5}\cmidrule(lr){6-7}\cmidrule(lr){8-9}\cmidrule(lr){10-11}
    Model &
      \multicolumn{1}{c}{Top} & \multicolumn{1}{c}{Mean} &
      \multicolumn{1}{c}{Top} & \multicolumn{1}{c}{Mean} &
      \multicolumn{1}{c}{Top} & \multicolumn{1}{c}{Mean} &
      \multicolumn{1}{c}{Top} & \multicolumn{1}{c}{Mean} &
      \multicolumn{1}{c}{Dev} &
      \multicolumn{1}{c}{Test}\\
    \midrule
    
    Linear& 2.56 & 2.56$\pm$0.& 1.58 & 1.58$\pm$0.& 11.9 & 11.9$\pm$0.& 9.25 & 9.25$\pm$0.& 6.62 & 4.96\\
    
    \abr{ir} & 9.48 & 9.48 & 6.23 & 6.23 & 62.2 & 62.2 & 54.5 & 54.5 & 45.8 & 38.8\\
    
    \abr{dan}& 10.7 & 10.4$\pm$0.3& 8.28 & 7.88$\pm$0.3& 60.0 & 59.1$\pm$0.9& 51.0 & 51.4$\pm$1& 42.6 & 35.5\\
    
    \abr{rnn}& 10.5 & 9.46$\pm$0.7& 7.86 & 7.78$\pm$0.4& 52.3 & 51.8$\pm$1& 46.4 & 45.9$\pm$0.9& 27.6 & 23.3\\
    
    \abr{bert}& 12.5 & 11.1$\pm$0.8& 9.34 & 9.49$\pm$0.3& 53.4 & 55.0$\pm$0.9& 47.0 & 48.8$\pm$0.9& 36.6 & 31.6\\

    \bottomrule
	\end{tabular}

%% file: 2021_jmlr_qanta/sections/70-live.tex
\section{Live Exhibition Events}
\label{sec:live}

No amount of thorough experimentation and analysis of machine learning
systems can match the public interest and rubber-meets-the-road
practicalities of live matches between humans and machines.
\abr{ibm}'s Watson in Jeopardy!~\citep{Ferrucci2010BuildingWA}, Deep
Blue in chess~\citep{Hsu1995DeepBS}, and Google's AlphaGo in
Go~\citep{Silver2016MasteringTG} were both tremendous scientific
achievements and cultural watersheds.
In the case of chess and Go, they transformed how the games are played
through insight gained from the collaboration of humans and machines.
Lastly, although our offline evaluation is reasonable, a live evaluation verifies that the two correspond since that is not always clear~\citep{hersh2000eval}.

In a similar spirit, we have hosted eight live events since 2015 where we showcase our research to the public by having humans and machines compete against each other.\footnote{
  Videos of our events are available at \url{http://events.qanta.org}.}
Except for our \abr{nips} 2015 Best Demonstration against \abr{ml} researchers, our system's opponents have been strong trivia players.
Their achievements include victories in numerous national \qb{} championships (high school and college), Jeopardy!, and similar trivia competitions.

Our inaugural event in 2015 at the \qb{} High School National Competition Tournament (\abr{hsnct}) pitted an early and vastly different version of our system against a team of tournament organizers in a match that ended in a tie.\footnote{
  Our software did not handle ties correctly and terminated instead of playing tiebreaker questions.}
Later that year, a similar system defeated Ken Jennings of Jeopardy! fame at the University of Washington, but lost convincingly (145--345) at \abr{hsnct} 2016.
The subsequent year at \abr{hsnct} 2017, our redesigned system narrowly defeated its opponents (260--215).
This system was used an \abr{ir} guesser with \abr{rnn} buzzer combined with a question type classifier~\citep{li2002class}.\footnote{
  In absolute terms, the type classifier did not improve accuracy; however, in our matches we display the top five scoring guesses and the type classifer improved the ``plausibility'' of that list.
}
Although this impressive result appears to follow the trend of machines improving until they defeat skilled humans, it is far from the whole story.

In parallel with these events, we hosted events where teams---humans
and machines---were selected from open competition.
Our first of these style events was hosted as part of a \abr{naacl}
2016 workshop on question answering.
Before the event local high school teams competed against each other,
and researchers submitted their machine systems which also played
simulated matches against each other.
At the event the best human and machine teams played against each
other with the high school team defeating an early version of Studio
\abr{ousia}'s
system~\citep{Yamada2017LearningDR,Yamada2018StudioOQ}.\footnote{ The
  \abr{ousia} system embeds words and entities separately, and uses a
  \dan{}-based architecture over these.}  In 2017, we hosted a similar
workshop at \abr{nips} where an improved version of \abr{ousia}'s
system yet again defeated its machine competition, but this time also
defeated the invited human team.

Events and collaborations like these show that \qb{} is more than just
another question answering task.
By engaging with the \qb{} community to digitize \qb{} questions in a
machine-readable form we not only made our research possible, but
started the ecosystem of tools that students now rely on to practice
with before competitions.
In the next step towards deeper collaboration with this community we
are building ways for humans and machines to cooperate in
competition~\citep{augment2018} and in writing
questions~\citep{wallace2018trick}.
We accomplish all this while simultaneously providing ways for
students of all ages to engage with and benefit from research through
our live exhibition events.

%% file: 2021_jmlr_qanta/sections/80-rel.tex
\section{Related Work}
\label{sec:rel}

Quizbowl is a question answering and sequential decision-making task.
This section begins by positioning \qb{}---as a \qa{} task---amongst early and more contemporary \qa{} and reading comprehension tasks including those that have shifted towards multi-step or adversarial questions.
Following this, we make move towards the decision-making aspect of \qb{} by first discussing connections with model calibration and prediction under domain shift.
Having discussed the uncertainty in model confidence scores, we finally discuss handling uncertainty caused by opponents through opponent modeling.

\subsection{Question Answering Datasets}
\label{sec:ds-tasks}

\qa{} and \abr{rc} datasets vary significantly in form and focus; here
we focus on factoid \qa{} and reading comprehension tasks.

The least complex types of questions are often called ``simple
questions'' since they can be answered by a single simple fact.
For example, SimpleQuestions~\citep{DBLP:journals/corr/BordesUCW15} as
specifically designed so that questions can be answered using one
knowledge-base triplet, and
WikiMovies~\citep{DBLP:conf/emnlp/MillerFDKBW16} automatically
generates questions from knowledge base triplets.
Similarly, WebQuestions~\citep{Berant2013SemanticPO} uses the Google
Suggest \abr{api} to collect questions containing specific entities
and crowdworkers only answer questions using Freebase facts.
Despite the relative ease in creating these datasets, they lack
complexity and linguistic diversity, leading near-complete solutions for
SimpleQuestions~\citep{Petrochuk2018SimpleQuestionsNS}.
In \qb{}, the capability of answering simple question like these is a
\emph{bare minimum requirement} and takes form as the final,
``giveaway'' clues at the end of questions which even novice (human)
players usually answer correctly.

Humans have played trivia games and tournaments for
decades~\citep{boydgraber2020nerds} and as a result there are ample
non-\qb{} sources of questions.
The most famous example---Jeopardy!---was converted into the
\searchqa{} dataset~\citep{Dunn2017SearchQAAN}.
Other trivia-based \qa{} datasets include TriviaQA which is built from
fourteen trivia sites~\citep{JoshiTriviaQA2017} and Quasar-T which is
built from questions collected by a reddit
user~\citep{Dhingra2017QuasarDF}.
While some of these are paired with potentially useful supporting
evidence, a hallmark of these datasets and \qb{} is that the question
alone unambiguously identifies an answer.

However, providing supporting documents to answer questions is another
popular way to frame \qa{} tasks.
The best examples of tasks with unambiguous questions and verified
supporting documents are \trecqa{}~\citep{Voorhees2000BuildingAQ} and
NaturalQuestions~\citep{nq19}.  In these tasks, the questions---user
queries from search engines---were written without knowledge of any
particular supporting documents and afterwards annotators attempted to
find appropriate supporting documents.
Although \triviaqa{} provides
potentially relevant documents, they are not human verified;
similarly, \abr{ms marco}~\citep{Nguyen2016MSMA},
WikiReading~\citep{DBLP:conf/acl/HewlettLJPFHKB16}, and
\newsqa{}~\citep{Trischler2017NewsQAAM} also provide unverified
supporting documents.
\squad{} in particular falls outside this
paradigm since---in general---questions are dependent on the selected
context paragraph.

Another set of tasks focuses on creating questions that require
multiple supporting documents and multi-step reasoning.
In \qb{}, early clues are often a composition of multiple facts about
the answer.
For example, to guess ``Die Zauberfl\"ote'' from ``At its premiere,
the librettist of this opera portrayed a character who asks for a
glass of wine with his dying wish'', one would have to combine two
pieces of text from Wikipedia: ``Papageno enters.  The priests grant
his request for a glass of wine and he expresses his desire for a
wife.''  and ``Emanuel Schikaneder, librettist of Die Zauberfl\"ote,
shown performing in the role of Papageno''.
While this work focuses on tossup \qb{} questions, a second type of
\qb{} question---bonuses---emphasize this multi-step aspect through
multi-part questions~\citep{Elgohary:Zhao:Boyd-Graber-2018}.

Multi-step reasoning datasets primarily differ by providing
ground-truth annotations to sufficient supporting documents.
In Wikihop~\citep{Welbl2018ConstructingDF}, multi-hop questions are
automatically constructed from Wikipedia, Wikidata, and WikiReading.
HotPotQA~\citep{Yang2018HotpotQAAD} follows a similar structure, but
question text is crowdsourced rather than automatically generated.
However, show \citet{Min2019CompositionalQD} that although these
questions are meant to be only solvable by using multi-step reasoning,
that many are answerable with single-hop reasoning.
Subsequent datasets like \abr{qasc}~\citep{Khot2019QASCAD},
\abr{drop}~\citep{Dua2019DROPAR}, and
\abr{break}~\citep{wolfson2020break} focus on creating questions that
are much more likely to require multi-step reasoning through
adversarial annotation.
However, multi-step questions are not the only way to make questions
more difficult.

Adversarial question authoring and adversarial
filtering~\citep{zellers2018swag} are other ways to increase
difficulty.
The general framework of adversarial authoring filters questions
either during or after annotation by whether or not a strong baseline
answers it correctly.
For example, \citet{wallace2018trick} show that when \qb{} question
writers---while authoring a question---are shown what a model would
answer and why, that they create questions that are significantly more
difficult for machines while being no more difficult for human
players.
Along similar lines, \citet{bartolo2020beat} let writers see what a
model would answer, but iteratively create new adversarial questions,
re-train the model, and collect new adversarial questions anew.
Although contrast sets explicitly do not have a model in the
loop~\citep{gardner2020contrast}, they are similarly intended to
challenge models through example perturbations.
Other effective example perturbations for automatically creating
adversarial examples include adding sentences to context
paragraphs~\citep{Jia2017AdversarialEF} and in general
semantic-preserving permutations~\citep{Singh2018SemanticallyEA}.
The common thread in all these works is to make questions more
difficult for machines so that models continue to improve.

\subsection{Answer Triggering and Model Calibration}
\label{sec:calibration}

Just as in \qb{}'s buzzer, in real-world applications of machine
learning it is important to know when to trust a model's confidence in
its predictions~\citep{jiang2012med}.
In \qa{}, knowing when to answer is known as answer
triggering~\citep{Yang2015WikiQAAC,Rajpurkar2018KnowWY} and the core
task---correctly estimating model confidence---is model
calibration~\citep{zadrozny2001calib}.
Having machines that accurately convey their confidence is doubly
important since humans have the unfortunate tendency to place too much
trust in machines~\citep{metzger2007digital}.
Despite this, the rise of deep learning seems to have only made this
more challenging~\citep{Guo:Pleiss:Sun:Weinberger-2017,feng2018rawr}.
Fortunately, recent work has made progress in this problem by making
connections to out-of-domain detection~\citep{kamath2020selective}.
Along similar lines, the buzzing task in \qb{} partially dependent on
having well-calibrated models.

\subsection{Opponent Modeling}

\qb{} is far from the only game where players benefit from modeling
opponent behavior.
Opponent modeling is particularly important in games with hidden
information---in \qb{} this takes the form of the yet-to-be-revealed
question and the per-question skill of the opponent.
One classic example of opponent modeling under uncertainty is
Poker~\citep{billings1998poker} where identifying and adapting to
opponents is central to the game.
In games like Scrabble, opponent behavior can even be used to infer
hidden information---such as their remaining tiles---to make it easier
to anticipate and counter their strategy~\citep{richards2007scrabble}.
Similarly, in real time strategy games the opponent's strategy is
often hidden and hierarchically structured~\citep{schadd2007rts}; they
may play aggressive---similar to aggressive buzzing---making defensive
play more advantageous.
Traditionally, \qb{} is played in teams so a full model should account
for your own team's skills as well as the mixture of opponent skills
(e.g., certain players may be better at history questions).
Although this is unaddressed in \qb{}, this general framing has been
considered in games where players compete with each other for limited
resources, but are advantaged by doing so
semi-cooperatively~\citep{osten2017minds}.
One potential area of future work is in focusing on opponent modeling
in \qb{} with an emphasis on accounting for teams of players and
strategies that evolve over the course of a match (e.g., play less
conservatively when trailing).

%% file: 2021_jmlr_qanta/sections/85-future.tex
\section{Future Work}
\label{sec:future-work}
Here we identify research directions that are particularly well suited to the \qanta{} dataset and \qb{}.
Rather than focus on general research directions in question answering we identify areas that best take advantage of \qb{}'s unique aspects.
Specifically we focus on directions using its interruptible and pyramidal nature, the perpetual influx of new and diverse questions from annual tournaments, and the supportive community interested in making \qb{} even more supportive of scholastic learning and achievement.

\subsection{Generalization in Factoid Question Answering}

\begin{figure}[t]
    \centering
    \includegraphics[width=0.8\textwidth]{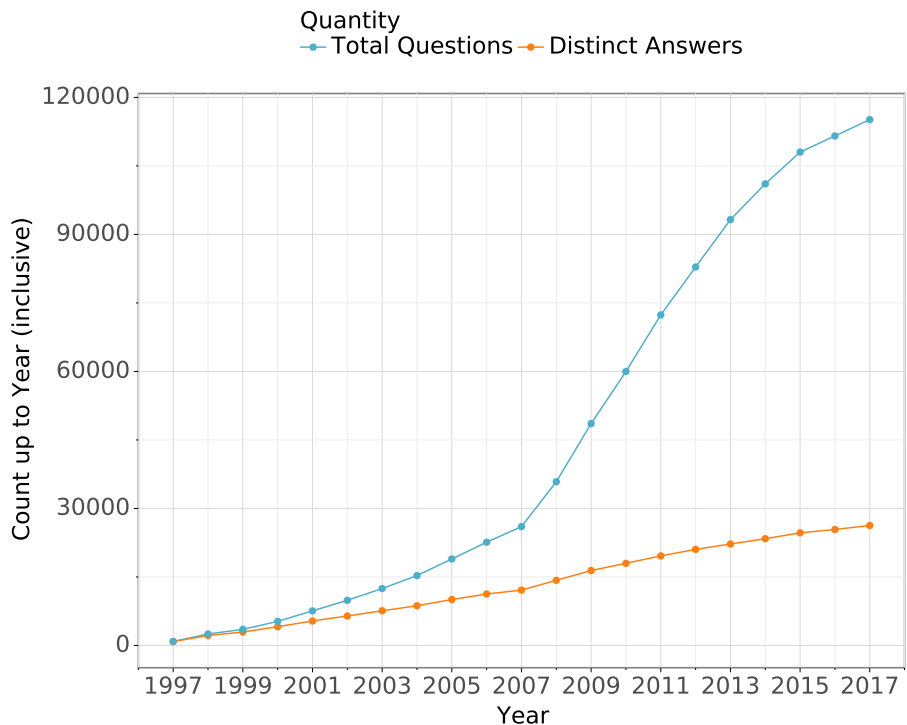}
    \caption{
        The growth of the \qanta{} dataset in number of questions and number of distinct answers over the past twenty years starting in 1997.
        The dataset has grown by at least 5,000 questions every year since 2010.
        All questions with matched answers are included, and we construct the plot by using the tournament year of each question.
        Independently, participation in \qb{} (and thus number of students writing questions) has roughly doubled every year since 2008.
    }
    \label{fig:size}
\end{figure}

Although some syntactic forms in \qb{} may be overly specific, its
ever-growing size and diversity makes it attractive for studying more
generalizable factoid question answering.
The reasons for this are twofold.
First, as discussed in Section~\ref{sec:datasets}, the \qanta{}
dataset is already diverse in topics, syntax, and in range of answers
(over twenty-five thousand considering all data folds).
Second and more importantly though the dataset is growing year over
year.  Since 2007 the size of the dataset has over quadrupled, and the
growth shows no sign of slowing down.
Figure~\ref{fig:size} shows this growth over the past twenty years.

As the dataset continues to grow it will demand that machines and
humans broaden their knowledge of past events while also updating
their knowledge with current events.
Every year presents an opportunity to test both how well models
generalize to novel questions and how well they generalize to
questions about current events.
For example, in a 2017 exhibition match our model missed a question
about the company responsible for driving down rocket launch costs
(SpaceX); a phenomenon which only manifested itself several years
prior.
With this ever present influx of new questions every year we believe
this opens new research directions in testing the generalization of
models.

\subsection{Few Shot Learning and Domain Adaptation}
One well-suited research direction with \qb{} is adopting methods from
domain adaptation and few-shot learning for factoid question
answering.
A large source of error in our models is a scarcity of training
examples for most answers (see Figures~\ref{fig:ntrain-cdf}
and~\ref{fig:acc_ntrain}).
Zero and few-shot learning in \abr{nlp} and Computer vision face
similar challenges of using shared structure to improve the use of the
scarce training data~\citep{Xian2018ZeroShotL}.
Correcting errors induced by the scarcity of training examples is an
exciting research direction.
From (frustratingly) easy methods like simple feature augmentation with source and target domain features~\citep{Daum2007FrustratinglyED,Kim2016FrustratinglyEN} to more sophisticated adversarial methods~\citep{multiDA2018}
there are a variety of options investigate.

\subsection{Robust, Explainable and Trustable Machine Learning}
\label{sec:augment}

Developing robust~\citep{goel2021gym}, explainable~\citep{Belinkov2019AnalysisMI} and trustable~\citep{mitchell2019model} machine learning systems overlaps
with the goals of machine learning researchers and the \qb{}
community.
For example, continually evolving, robustness-centric evaluations\footnote{Platforms like \url{https://dynabench.org} and \url{https://robustnessgym.com} propose frameworks for improving robustness evaluations.}
would pair well with the annual writing of new questions.
A concrete direction for future work is to integrate adversarial question authoring~\citep{wallace2018trick,bartolo2020beat} into these dynamic evaluations.
While these works focus on creating explicitly adversarial examples, an alternative approach to creating difficult questions is to expand diversity of questions and answers since they are naturally difficult for \qa{} models (Section~\ref{sec:divergence}).
Ideally, \nlp{} models should eventually be robust to adversarial and diverse inputs, but that likely requires improving \emph{both} data and models.

Humans can improve at games like Chess and Go by learning from
machines; similarly, if models are explainable \qb{} players can learn from and cooperate with
machines as well.
For example, \citet{augment2018} built and evaluated interpretations
of machine learning models based on how effective they were at
improving human live play.
A related research direction would be to use these interfaces and
insights about models of human learning---such as the effectiveness of
spaced repetition~\citep{ebbinghaus}---to improve knowledge retention
as \citet{Settles2016ATS} do for language learning.
Both of these directions fuse work in human-computer interaction and
interpretation of machine learning algorithms.

Our collaborative research in \qb{} thus far is only a beginning.
\qb{} supports work in factoid question answering and sequential
decision-making.
We list several challenges in these based on our experiments, but
there are certainly more.
Beyond playing \qb{}, there are opportunities in human-in-the-loop
research that can improve interpretations of machine learning models
while producing useful artifacts such as adversarial datasets.
We hope that our work in establishing the \qanta{} datasets and \qb{}
as a machine learning task empowers others to contribute to these and
other future research.

%% file: 2021_jmlr_qanta/sections/90-conc.tex
\section{Conclusion}
\label{sec:conc}

This article introduces and argues for \qb{}: an incremental question answering task.
Solving \qb{} questions requires sophisticated \abr{nlp} such as resolving complex coreference, multi-hop reasoning, and understanding the relationships between the plethora of entities that could be answers.
Fundamental to answering \qb{} questions is that the questions are incremental; this is both fun and good for research.
It is fun because it allows for live, engaging competitions between humans and computers.
This format---the product of refining human question answering competitions over decades---is also good for research because it allows for fair, comprehensive comparison of systems and iterative improvement as systems answer questions earlier and earlier.

To evaluate systems we use three methods: offline accuracy-based metrics adapted to the incremental nature of \qb{}, simulated matches against machines and humans, and live exhibition matches.
Although the best models have sixty percent accuracy at the end of questions, this is well below the best players, and the first clues remain particularly challenging.

Improving \qb{} models can incorporate many commonplace tasks in \abr{nlp} other than question answering.
Reasoning about entities can be improved through better named entity recognition, entity linking, coreference and anaphora resolution.
Some of the more difficult clues in \qb{} however presume that the player has read and integrated the content of books such as important plot points.
Further work in reading comprehension and summarization could help answer some of these questions.
At a more general level, the extraction of information from external knowledge sources (such as books or Wikipedia) is important since the distribution of training examples per answer is heavily skewed and some new questions ask about current events.
Improving \qb{} models requires and can further motivate advances in these tasks and others.

However, the benefits to research go beyond format or specific sub-tasks, and extend to our symbiotic collaboration with the public.
Exhibition matches double as outreach events and opportunities to put machine systems to the test on previously unseen questions.
Another area of active research is in collaborating with the \qb{} community to further improve the quality of questions for humans and machines alike.
Writers are empowered with machine learning tools to discover bad clues which helps create questions more interesting to humans that consequently better test the generalization of systems.
In this the goals of the \qb{} and communities align; we both seek to create datasets that discriminate different levels of language and knowledge understanding.

Beyond the specific \qb{} community, live exhibitions also serve the general public.
Exhibition games demonstrate what \abr{nlp} and \abr{ml} systems can do and what they cannot.
When the tricks that computers use to answer question are revealed to lay audiences, some of the mystique is lost, but it can also encourage enthusiasts to investigate our techniques to see if they can do better.
Our open data and code facilitates this open competition, and the \qb{} community helps make it fun and engaging.

\qb{} isn't just another dataset or task; it is a rich platform for \abr{nlp} research that co-evolves with the \qb{} community.
From the digitization of \qb{} questions to our online interface which created and popularized online \qb{} play.
From cooperative play between humans and machines to providing tools for better writing better question we have built a symbiotic relationship that continues to yield productive research while moving the \qb{} community forward.
We hope that new unforeseen research directions will continue emerging from this collaboration while simultaneously giving back to the \qb{} community through new and exciting ways of engaging with state-of-the-art research in machine learning and natural language processing.

%% file: 2021_jmlr_qanta/sections/appendix.tex
\section{Preprocessing}
\label{apx:preprocess}

This section provides a detailed description of preprocessing on the question dataset.

\subsection{Aligning and De-duplicating Questions}
Since we obtain machine-readable versions of questions from two online sources it is necessary to ensure that we do not include the same question twice.
We use the metadata associated with each question such as tournament and year.
As part of our preprocessing we manually align the values of these fields.\footnote{
  We also align category and sub-category fields.
}
We use these fields to ensure that questions for each tournament and year are included only once.

\subsection{Textual Preprocessing}
Models should define their own textual preprocessing so we only preprocess the text to remove \qb{} specific artifacts.
Most of these artifacts are instructions to the moderator or organizer such as ``MODERATOR NOTE:'', ``Description Required'', ``15 pts:'', or a reference to the category of the question; we use regular expression rules to remove these.
Since we report results on accuracy after the first sentence in questions we also provide a set of canonical sentence tokenization indices computed using spacy.\footnote{\url{https://spacy.io}}

\subsection{Fold Assignment}
\label{apx:f-assign}
We divide \qanta{} dataset questions into training, development, and test
folds based on the competitiveness and year of the source tournament. Since
championship tournaments typically have the highest quality questions we use
questions from championship tournaments 2015 and onward as development and
test sets. All other questions are used as the training set.

Table~\ref{table:dataset-folds} shows the divisions of each fold; each
train, dev, or test fold is assigned to be used for either determining
\emph{what} to answer (guessing) or \emph{when} to answer (buzzing).  Questions
in ``guess'' folds are used for developing question answering systems as in
Section~\ref{sec:guessing}. Questions in the ``buzz'' folds are used for
developing agents that decide when to answer as in Section~\ref{sec:buzzing}.

When we assign folds to \qb{} questions we aim to create useful splits for
guessing and buzzing while preserving the integrity of the development and test
sets. Namely, when we create test and development folds we make the division
into folds not depend on whether or not gameplay data exists. If it were the
case that by making this unconditional assignment the number of questions with
gameplay data is too small this would be a problem. We do not find this
to be a problem however.

For test set questions this is easily accomplished by using an implicit
quality filter and a temporal split; use only questions from national
championship tournaments, questions from 2016 are used in the buzzing test set,
and questions from 2017 and 2018 are used for the guessing test set. Following
this we pair the test fold for buzzing with gameplay data, and are fortunate
that the number of questions is not small.

To create the development sets we use questions from 2015 which are randomly
split with equal probability into guessing and buzzing specific folds.
Similarly to the test set we associate gameplay data after this assignment
occurs to preserve its integrity against any bias that conditioning on having
gameplay data would have.

For the training data we make a weaker attempt to eliminate bias in favor of
ensuring that the training folds for guessing and buzzing are large enough. We
first divide the training questions with an 80/20 split. Questions in the
eighty percent split are assigned to the guessing fold. Each remaining question
is assigned to the buzzing fold if it has gameplay data, otherwise it is
assigned to the guessing fold. Figure~\ref{table:dataset-folds} shows the
result of this folding procedure.

\subsection{Matching QB Answers to Wikipedia Pages}
\label{apx:matching}

The automatic rule based part of this process is composed of two phases: an
expansion phase in that produces variants of the answer text, and a match phase
that determines when one of these variants is a match to a Wikipedia page. The
rules in the expansion phase can be as simple as exact text match to expanding
``The \{Master of Flémalle\} or Robert \{Campin\}'' to ``\{Master of
Flémalle\}'' and ``Robert \{Campin\}''. In this case, multiple
matches result in ``Robert Campin'' being the answer page: after
removing braces ``The Master of Flémalle'' Wikipedia redirects to ``Robert
Campin'' and ``Robert Campin'' is also an exact match.
These rules are incredibly effective at finding answers buried in \qb{} specific notation such as the random sample in Table~\ref{table:answer-strings}.
When matches disagree
we use the match that modified the original answer text the least.

\begin{table*}[t]
  \begin{center}
    \begin{tabularx}{\textwidth}{ X l }
      \toprule
      Original \qb{} Answer                                                       & Matched Wikipedia Page                    \\
      \toprule
      Nora Helmer                                                                 & A\_Doll's\_House                          \\ \bottomrule
      \{Gauss\}'s law for the electric field                                      & \textbf{No Mapping Found}                 \\ \bottomrule
      Thomas Hutchinson                                                           & Thomas\_Hutchinson\_(governor)            \\ \bottomrule
      linearity                                                                   & Linearity                                 \\ \bottomrule
      \{caldera\}s                                                                & Caldera                                   \\ \bottomrule
      William Holman \{Hunt\}                                                     & William\_Holman\_Hunt                     \\ \bottomrule
      \{plasma\}s                                                                 & Plasma\_(physics)                         \\ \bottomrule
      \{Second Vatican Council\} [or \{Vatican II\}]                              & Second\_Vatican\_Council                  \\ \bottomrule
      \{Jainism\}                                                                 & Jainism                                   \\ \bottomrule
      \{Electronegativity\}                                                       & Electronegativity                         \\ \bottomrule
      Hubert Selby, Jr.                                                           & Hubert\_Selby\_Jr.                        \\ \bottomrule
      (The) Entry of Christ into Brussels (accept equivalents due to translation) & Christ's\_Entry\_Into\_Brussels\_in\_1889 \\ \bottomrule
      Depictions of Speech [accept equivalents]                                   & \textbf{No Mapping Found}                 \\ \bottomrule
      stress                                                                      & Stress\_(mechanics)                       \\ \bottomrule
    \end{tabularx}
  \end{center}
  \caption{
    A random sample of \qb{} answer strings and their matched Wikipedia pages.
    Answer mappings are easy to obtain accurately since most failures in exact matching are due to \qb{} specific syntax that can be accounted for by rule based matching.
    Combined with manual annotation to find common non-exact matches, this process succeeds on \nquestions{} of \ntotalquestions{}.
  }
  \label{table:answer-strings}
\end{table*}

There are inevitably cases where the automatic system fails to find a match, or
finds the wrong match. Qualitatively these are often caused by
disambiguation errors such as failing to differentiate between ``Guernica'' the
city versus the painting by Picasso, small differences in answer strings, and
when there is no suitable Wikipedia page. To correct or verify these errors we
(the authors), and skilled members of the QB community (such as tournament
organizers and participants from our exhibition matches) manually annotated a
significant fraction of the training data, and all the test data.

Rather than doing manual annotation of each question, we begin by defining
mappings of answer strings to Wikipedia pages so that when that string occurs
multiple times it does not require manual annotation for every occurrence of
that answer in questions. However, this has the serious drawback that if the
answer string is ambiguous then it may result in mislabeled answers. To avoid
this problem we design a manual process whereby annotators update three sets of
answer-to-Wikipedia mappings: unambiguous, ambiguous, and direct mappings.

Unambiguous annotations contain a list of answer strings that when seen map to
a specific Wikipedia page. As the name implies, we only insert annotations here
when the answer unambiguously identifies the corresponding Wikipedia page.
Ambiguous annotations similarly contain a list of answer strings, but are
paired with a list of disambiguation words. If the answer string is seen, at
least one word is in the question text, and there are no other ambiguous
matches, then it is mapped. For example, if the answer string is ``amazon''
and the question contains the word ``river'' then we assume ``Amazon river'' is
the correct page while if the question mentions ``bezos'' then the correct page
is ``Amazon (company)''. Finally, direct mappings match the answer for specific
questions.

The last major design decision in this process addresses how we prevent
information from the test data to leak into the training data. The root of the
data leak issue is that the distribution of answers between training and test
data often results in only approximately 80\% of test set answers occurring in
the training data. We observed this phenomena empirically in both our data and
the distribution of answers from our numerous exhibition events. If all answer
strings are naively combined, then mapped, this implies that the training data
will be biased towards its answers containing an over abundance of test set
answers.
A major difference between this and prior versions of the \qanta{} dataset is finding and fixing this issue.

We correct this error by separating the answer string pool for training and
test questions. Although this results in more annotation work, it avoids
information leakage. While reviewing our annotation procedure we noticed
another source of bias. Recall that we do not exhaustively annotate the
training data. In our initial annotation we did not fully annotate the test
data, and by doing so introduced a bias towards easier-to-annotate questions in
the test set. To eliminate this bias---and make it as similar to playing a
\qb{} tournament as possible---we annotated every question in the test
set.\footnote{Specifically, we either pair each test set answer strings with a
  Wikipedia title or mark it as not having a corresponding Wikipedia title.}



\subsection{Buzzer features}
\label{apx:buzzer_features}
The guesser updates its list of guesses whenever a new word of the \qb{}
question is revealed. At each time step, the buzzer extracts features from both
the current and all past guesses and predict whether the current guess is
correct.
It is important to include past guesses as the dynamics of the guesser's
confidence contains strong signal about its correctness: the guesser
usually starts with some random guess when little information is provided, then
fluctuates between several plausible answers---just as humans do, and finally
stablizes to a single answer, at which point the buzzer should buzz.
Below is the full list of buzzer features we use in the experiments:
\begin{itemize}
  \item Probabilities of the current top 3 guesses
  \item Change of top 3 probabilites from the previous step
  \item Gaps between probabilities of the top 3 guesses
  \item Binary indicator of whether each of the top 5 guesses increased its
        ranking from previous step
  \item Mean and variance of probabilities of the current top 3 guesses
  \item Mean and variance of probabilities of the previous top 3 guesses
\end{itemize}

\section{Natural Questions Categories}
\label{apx:nq}

Section~\ref{sec:topical-div} analyzes the topical diversity of \qb{} questions and makes comparisons to NaturalQuestions.
To compare to NaturalQuestions---which does not have category labels---we annotated a random subset of $150$ questions using \qb{}'s categories (Table~\ref{table:nq-annot}).

\begin{table}
  \centering
  \begin{tabular}{ l r r}
    \toprule
    Category          & N    & Percent  \\
    \midrule
    Pop Culture       & $55$ & $40\%$   \\
    History           & $26$ & $19\%$   \\
    Science           & $20$ & $15\%$   \\
    Other             & $13$ & $9.6\%$  \\
    Social Science    & $7$  & $5.1\%$  \\
    Geography         & $6$  & $4.4\%$  \\
    Religion          & $2$  & $1.5\%$  \\
    Literature        & $5$  & $3.7\%$  \\
    Philosophy        & $1$  & $0.74\%$ \\
    Fine Arts         & $1$  & $0.74\%$ \\
    \midrule
    Total w/ Category & 136  & $100\%$  \\
    \midrule
    No Category       & 14              \\
    \midrule
    Total             & 150             \\
    \bottomrule
  \end{tabular}
  \caption{
    A breakdown of NaturalQuestion example topics using \qb{} categories.
    Most questions are about pop culture and the distribution includes many fewer questions about Literature and Fine Arts.
  }
  \label{table:nq-annot}

\end{table}